\documentclass{article} 
\usepackage{iclr2027_conference,times}

\usepackage{amsmath,amsfonts,bm}

\def\eqref#1{equation~\ref{#1}}
\def\Eqref#1{Equation~\ref{#1}}

\def\1{\bm{1}}

\DeclareMathAlphabet{\mathsfit}{\encodingdefault}{\sfdefault}{m}{sl}
\SetMathAlphabet{\mathsfit}{bold}{\encodingdefault}{\sfdefault}{bx}{n}

\DeclareMathOperator*{\argmax}{arg\,max}

\usepackage[colorlinks = True, linkcolor = violet, citecolor = blue!50!white]{hyperref}
\usepackage{url}
\usepackage{algpseudocode}
\usepackage{algorithm}
\usepackage{amssymb}
\usepackage{enumitem}
\usepackage{graphicx}
\usepackage{subcaption}
\usepackage{booktabs}
\usepackage{pifont}
\usepackage{amssymb}
\usepackage[dvipsnames]{xcolor}
\usepackage{adjustbox}

\newcommand{\cmark}{\textcolor{green!60!black}{\ding{51}}}
\newcommand{\xmark}{\textcolor{red!70!black}{\ding{55}}}

\title{In-Context Batch Black-box Optimization via Flow Matching}

\iclrfinalcopy

\author{Quoc-Anh Hoang Nguyen\textsuperscript{1,2}, Xinyu Zhang\textsuperscript{1,2}, Daolang Huang\textsuperscript{1,2}\\
\textbf{Julien Martinelli\textsuperscript{*,1,2}, Samuel Kaski\textsuperscript{*,1,2,3}} \\
\textsuperscript{1}ELLIS Institute Finland \\
\textsuperscript{2}Department of Computer Science, Aalto University, Finland \\
\textsuperscript{3}Department of Computer Science, University of Manchester \\
\textsuperscript{*}Equal contribution
}

\begin{document}

\maketitle

\begin{abstract}

Batch black-box optimization is important for expensive systems where many candidates can be evaluated in parallel, yet only a few rounds of evaluation are affordable.
In this regime, effective batch construction must identify multiple promising regions while avoiding redundant evaluations, and conventional Bayesian optimization can require increasingly difficult batch-selection procedures as the batch size grows.
We introduce Flow Matching for Thompson-Sampling Refinement (FMTR), an amortized approach that learns offline a continuous, history-conditioned proposal distribution from which variable-size batches can be sampled. FMTR is trained offline by conditional flow matching on Thompson-sampling argmaxes refined toward high-value regions via a horizon-dependent Langevin pushforward map, adapting the exploration-exploitation tradeoff over the optimization horizon.
At test time, the model parameters remain fixed and batch generation requires neither task-specific surrogate refitting nor numerical acquisition optimization. FMTR achieves strong performance against batch BO and in-context baselines on synthetic and real-world benchmarks, while substantially reducing the computational cost of batch selection.

\end{abstract}
\vspace{-.5cm}
\section{Introduction}

Many scientific and engineering design problems require optimizing an objective that can only be queried through costly experiments or simulations.
Modern experimental and computational platforms increasingly support parallel evaluation, from high-throughput laboratories to compute clusters, motivating batch black-box optimization in applications such as enzyme engineering, regulatory RNA design, and materials design~\citep{canty2025science,torres2026generative,batchbobio,batchbomaterial}.
This changes the nature of the optimization decision: in the batch setting, the utility of each candidate depends on the rest of the batch, which should cover several plausible high-value regions and underexplored parts of the search space while avoiding redundant evaluations. Moreover, the evaluation of each batch can incur substantial experimental or computational costs, making the number of optimization rounds an important practical constraint. Consequently, it is desirable to develop batch selection methods that make effective use of each round and achieve high-quality solutions with as few evaluation rounds as possible.

Bayesian optimization (BO;~\citealt{garnett2023bayesian}) is a standard approach for constructing such batches under limited evaluation budgets. It models the objective with a probabilistic surrogate, commonly a Gaussian process (GP,~\citealt{RasmussenW06}), and selects evaluations through an acquisition function that balances promising predicted values against posterior uncertainty.
However, when selecting $q$ candidates at once, the acquisition must account for interactions among the proposed points, making its maximization a joint problem over $qd$ variables.
 Joint acquisitions become difficult to optimize as $q$ grows, while common alternatives construct the batch through $q$ sequential or independent optimization problems~\citep{wilson2018maximizing,kandasamyparallelts,balandat2020botorch,teufel2024}. Each round thus requires updating a task-specific surrogate and solving a difficult, time-consuming batch-selection problem before the evaluations can be run in parallel.

In-context learning offers a complementary route by moving recurring computation to offline pretraining over a distribution of optimization tasks. At deployment, a pretrained model adapts to a new objective through the observations supplied in context, without updating its parameters. Existing methods introduce in-context learning at different stages of the BO loop. Surrogate-based approaches perform posterior inference in context but retain a conventional acquisition function and an online search over the design space~\citep{muller2023pfns4bo,pmlr-v258-chang25a,yu2026git}. Acquisition-level approaches instead learn to predict acquisition values, replacing their potentially expensive computation while still searching for the maximizing candidate~\citep{hung2025boformer,rakotoarison2026}. End-to-end in-context optimizers go further by learning the query policy itself, commonly from trajectory-level rewards, and map the observed history and a supplied candidate set directly to the next query~\citep{maraval2023endtoend,zhang2025pabbo,zhang2026incontext,huang2025aline}. These policies remove task-specific surrogate fitting and numerical acquisition optimization, but are typically formulated as single-query policies over a finite candidate set. They therefore do not directly provide an in-context sampler over continuous query locations from which a batch can be drawn.

Generative models offer a natural way forward. By learning a proposal distribution over continuous designs, they turn batch construction into sampling: the proposal places mass on several promising regions, generating a batch of the desired size without optimizing a joint acquisition.
Recent generative BO methods either adapt a diffusion sampler using a task-specific surrogate or fit the proposal directly from observed utilities~\citep{yun2025posterior,oliveira2026generative}.
In both cases, the proposal is fitted or updated on the current objective as optimization proceeds, rather than produced in context by a model pretrained across tasks. This leaves open how to combine in-context adaptation with a reusable continuous proposal and how to specify its target distribution for each observed history.

\begin{figure}[h!]
    \centering
    \includegraphics[width=0.8\linewidth]{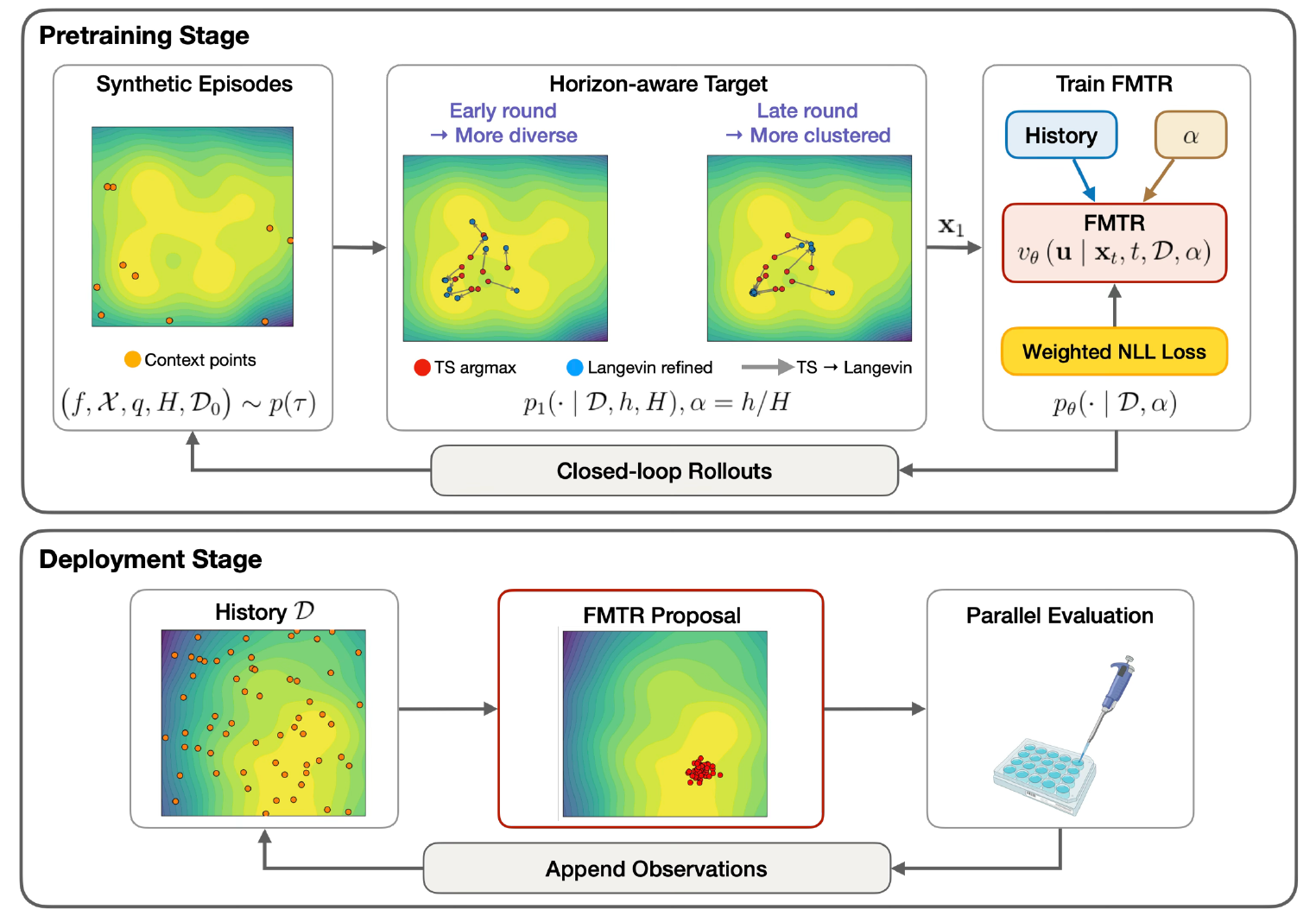}

    \small
    \setlength{\tabcolsep}{5pt}
    \begin{adjustbox}{max width=\linewidth}
    \begin{tabular}{@{}lccc@{}}
        \toprule
        Method &
        \shortstack{In-context\\amortization} &
        \shortstack{Batch\\selection} &
        \shortstack{Learned query\\distribution} \\
        \midrule
        Batch BO~\citep{kandasamyparallelts,teufel2024}
            & \xmark & \cmark & \xmark \\
        ICL-based BO~\citep{rakotoarison2026,yu2026git}
            & \textbf{\textcolor{orange}{Surrogate or acqf only}} & \xmark & \xmark \\
        End-to-end ICL optimization~\citep{maraval2023endtoend,zhang2026incontext}
            & \cmark & \xmark & \xmark \\
        Generative BO~\citep{yun2025posterior,oliveira2026generative}
            & \xmark & \cmark & \cmark \\
        \textbf{FMTR (ours)}
            & \cmark & \cmark & \cmark \\
        \bottomrule
    \end{tabular}
    \end{adjustbox}

\caption{\textbf{Overview of FMTR and its positioning relative to related methods.}
FMTR directly models a query distribution for end-to-end batch selection. \textbf{(Top)} Pretraining and deployment with closed-loop rollouts for data generation. }
    \label{fig:placeholder}
\end{figure}

\textbf{Contributions.}
We address this gap with \emph{\textbf{F}low \textbf{M}atching for \textbf{T}hompson-Sampling \textbf{R}efinement} (FMTR), which combines in-context adaptation with sample-based generation for large-batch-few-round black-box optimization tasks. Figure~\ref{fig:placeholder} provides an overview of our method and positions it against concurrent baselines. Our contributions can be summarized as follows:
\begin{itemize}[leftmargin=*]
\item We introduce a conditional flow-matching model that represents a proposal distribution over continuous query locations, conditioned on the observed history and remaining evaluation budget. FMTR generates a batch of the requested size by sampling from this proposal with fixed model parameters, without task-specific surrogate fitting or acquisition-function optimization.

\item We construct supervised training targets by refining Thompson-sampling argmaxes toward high-value regions according to the remaining optimization horizon. This provides a simple alternative to trajectory-level reinforcement learning while retaining more exploration early and increasing exploitation toward the end of optimization.

\item Across synthetic and real-world benchmarks, FMTR achieves strong optimization performance at substantially lower batch-selection cost. Ablations further show that target refinement and value weighting both improve over directly amortizing parallel Thompson sampling.

\end{itemize}

\section{Background}

We consider an expensive black-box objective \(f:\mathcal{X}\rightarrow\mathbb{R}\) over a search space \(\mathcal{X}\subset\mathbb{R}^{d}\), with
\begin{equation}
    \mathbf{x}^{\star}\in\argmax_{\mathbf{x}\in\mathcal{X}} f(\mathbf{x}).
    \label{eq:bbo_objective}
\end{equation}
Starting from an initial history \(\mathcal{D}_{0}\), at round \(r=1,\ldots,H\), the optimizer selects a batch \(B_{r}=\{\mathbf{x}_{r,i}\}_{i=1}^{q}\subset\mathcal{X}\) conditioned on \(\mathcal{D}_{r-1}\). The \(q\) points are evaluated in parallel, yielding observations \(y_{r,i}=f(\mathbf{x}_{r,i})+\epsilon_{r,i}\), with \(\epsilon_{r,i}\overset{\mathrm{i.i.d.}}{\sim}\mathcal{N}(0,\sigma^{2})\). The history is then updated: \(\mathcal{D}_{r}=\mathcal{D}_{r-1}\cup\{(\mathbf{x}_{r,i},y_{r,i})\}_{i=1}^{q}\). We focus on the large-batch, few-round regime, where \(q\) is large and the batch horizon \(H\) is small.

\paragraph{Thompson sampling with Gaussian processes.}

A simple way to construct diverse batches is to sample plausible objective functions given the current observations and optimize each of them independently.
A Gaussian process (GP) prior over \(f\) induces a posterior
\(p(f\mid\mathcal{D})\) after conditioning on the observed history
\(\mathcal{D}\)~\citep{RasmussenW06}. Thompson sampling (TS) draws a function
from this posterior and evaluates its maximizer:
\begin{equation}
    \widetilde{f}\sim p(f\mid\mathcal{D}),
    \qquad
    \mathbf{x}_{\mathrm{TS}}
    \in
    \arg\max_{\mathbf{x}\in\mathcal{X}}
    \widetilde{f}(\mathbf{x}).
    \label{eq:ts_query}
\end{equation}
The posterior draw therefore induces a distribution over optimizer locations,
\begin{equation}
    p_{\mathrm{TS}}(\cdot\mid\mathcal{D})
    =
    \mathcal{L}\!\left(
        \mathbf{x}_{\mathrm{TS}}
        \,\middle|\,
        \widetilde{f}\sim p(f\mid\mathcal{D})
    \right),
    \label{eq:ts_argmax_distribution}
\end{equation}
where \(\mathcal{L}\) denotes the law of a random variable. Different
posterior paths may place their maximizers in different regions that remain
plausible under the posterior. Parallel TS forms a batch of
size \(q\) by drawing and maximizing \(q\) posterior paths independently,
allowing these optimizations to be performed concurrently~\citep{kandasamyparallelts}.

\paragraph{Flow Matching.}
Flow Matching (FM) is a class of generative models that transform a prior distribution $p_0$ to a target distribution $p_1$ by learning a time-dependent velocity vector field $v_t$. This vector field constructs a flow $\phi_t$ via the ordinary differential equation (ODE):
\begin{equation}
    \frac{d}{dt}\phi_t(\mathbf{x}_0)= v_t(\phi_t(\mathbf{x}_0)), \qquad  \phi_0(\mathbf{x}_0)= \mathbf{x}_0
\end{equation}
If $\mathbf{x}_0 \sim p_0$, $\phi_t$ acts as a push-forward map, creating distribution $p_t$ at each time $t$, e.g $p_t = [\phi_t]_*p_0$. For training, we construct a conditional flow and optimize the Conditional Flow Matching (CFM) objective. One of the most natural choices for conditional flow is a straight line from source point $\mathbf{x}_0$ to target point $\mathbf{x}_1$, where $\mathbf{x}_t=(1-t)\mathbf{x}_0+t\mathbf{x}_1$ for $t\in[0,1]$. Then, the CFM loss takes the form:
\begin{equation} \label{eq:CFM_loss}
    \mathcal{L}_{CFM}(\theta)=\mathbb{E}_{t,\mathbf{x}_1 \sim p_1, \mathbf{x}_0 \sim p_0}\left[\| v_{\theta}(\mathbf{x}_t,t)-(\mathbf{x}_1-\mathbf{x}_0)\|^2\right]
\end{equation}
Usually, $t \sim U(0,1)$ and $p_0$ follows a standard Normal distribution. At inference time, one can randomly sample from source distribution $\mathbf{x}_0 \sim p_0$ and generate a sample from target distribution by solving ODE from $t=0$ to $t=1$ using learned vector field $v_{\theta}$.

\section{Methodology}

We propose FMTR, which learns a conditional proposal
over continuous query locations, and samples batches in parallel. 
Section~\ref{sec:pretraining_tasks} defines the synthetic episodes and histories used for pretraining; Section~\ref{sec:target_distribution} constructs the query target distribution; Section~\ref{sec:model} introduces the conditional flow-matching model and training objective; and Section~\ref{sec:closed_loop_training} describes the closed-loop rollouts used to collect deployment-like histories. Model architecture is given in Figure \ref{fig:model_architecture}.

\subsection{Pretraining tasks and histories} \label{sec:pretraining_tasks}
\vspace{-.2cm}
We sample synthetic optimization episodes
\begin{equation}
\tau = \bigl( f_{\tau}, \mathcal{X}_{\tau}, q_{\tau}, H_{\tau}, \mathcal{D}_{0}^{\tau} \bigr) \sim p(\tau),
\end{equation}
where \(f_{\tau}:\mathcal{X}_{\tau}\subset\mathbb{R}^{d_{\tau}}\rightarrow\mathbb{R}\) is the objective, \(q_{\tau}\) is the batch size, \(H_{\tau}\) is the number of optimization rounds, and \(\mathcal{D}_{0}^{\tau}\) is an initial history containing \(n_{0,\tau}\) random evaluations.
At round \(r\in\{0,\ldots,H_{\tau}-1\}\), the model acts on the optimization state 
\vspace{-.15cm}
\begin{equation} s_{r}^{\tau} = \bigl(\mathcal{D}_{r}^{\tau},h_{r}^{\tau},H_{\tau}\bigr), \qquad h_{r}^{\tau}=H_{\tau}-r.
\end{equation} The history \(\mathcal{D}_{r}^{\tau}\) serves as context, while the supervised targets are continuous query locations sampled from the target distribution introduced in Section~\ref{sec:target_distribution}. 
Thus, a pretraining example is determined by both the sampled objective and a history encountered while optimizing it.

\textbf{Synthetic objective prior.} We construct each objective as
\begin{equation} f_{\tau}(\mathbf{x}) = f_{\mathrm{RFF},\tau}(\mathbf{x}) + f_{\mathrm{trend},\tau}(\mathbf{x}).
\label{eq:synthetic_objective}
\end{equation}
The first component is a differentiable random Fourier feature approximation to a GP sample path~\citep{rahimi2007random}, with kernel family and hyperparameters randomized across episodes. Importantly, its explicit form allows function values and gradients to be evaluated at arbitrary locations and retains the generating kernel required by the target construction in Section~\ref{sec:target_distribution}.
Next, $f_{\mathrm{trend}}$ adds a randomized quadratic trend, providing smooth global structure on top of the local GP variation. This yields optimization landscapes with more clearly defined basins and optima, a technique employed in related amortized optimization methods~\citep{yang2023mongoose,pmlr-v258-chang25a,zhang2025pabbo}.
The full RFF parameterization, trend definition, and sampling distributions are given in Appendices~\ref{app:pretraining_prior} and~\ref{app:kernel}. Figure~\ref{fig:training_samples} displays synthetic samples.

\textbf{Optimization histories.} Starting from \(\mathcal{D}_{0}^{\tau}\), we generate a trajectory by repeatedly proposing a batch, evaluating \(f_{\tau}\), and appending the resulting observations to the history. Thus, a single episode contributes training states with different context sizes and remaining horizons. Since the proposal model induces the relevant history distribution, we periodically regenerate these trajectories using a mixture of the target distribution and the current model proposal, as described in Section~\ref{sec:closed_loop_training}.

\vspace{-.2cm}
\subsection{Query target distribution} \label{sec:target_distribution}

For each synthetic episode \(\tau\) and optimization state
\((\mathcal{D},h,H)\), we construct a target distribution
\(p_{1,\tau}(\cdot\mid\mathcal{D},h,H)\) over query locations. Because the
episode is generated during pretraining, both the objective \(f_\tau\) and
the kernel \(k_\tau\) used to generate its GP component are known. This
information is used only to construct offline training targets: the kernel
defines a distribution over plausible optimizer locations given the current
history, while the objective is used to refine these locations according to
the number of rounds remaining. We omit the dependence of
\(p_{\mathrm{TS}}\) and \(p_1\) on \(\tau\) below.

\textbf{Thompson-sampling base distribution.}
We instantiate TS using the kernel \(k_\tau\) and trend
\(f_{\mathrm{trend},\tau}\) sampled for the episode:
\vspace{-.3cm}
\begin{equation}
p_{\mathrm{TS}}(\cdot\mid\mathcal{D}) = \mathcal{L}\!\left( \arg\max_{\mathbf{x}\in\mathcal{X}_{\tau}} \widetilde{f}_{\tau}(\mathbf{x}) \,\middle|\, \widetilde{f}_{\tau} \sim p_{\tau}(f\mid\mathcal{D}) \right).
\label{eq:task_ts_distribution}
\end{equation}
Under the GP posterior induced by the sampled task and current history, \(p_{\mathrm{TS}}\) represents uncertainty
over the optimizer location and can assign probability mass to several
regions that remain plausible optima. We construct posterior paths by
combining differentiable RFF prior draws~\citep{rahimi2007random} with the
pathwise update given by Matheron's rule~\citep{wilson2020efficiently}.
Because \(k_\tau\) and its hyperparameters are sampled as part of the
episode, these paths can be constructed without estimating kernel
parameters from the typically sparse history. Each path is maximized
independently using gradient-based optimization. Appendix~\ref{app:pathwise_posterior}
gives the complete construction.

\textbf{Refinement based on remaining rounds.}
TS conditions on the observed history but does not
explicitly account for the remaining horizon. To favor exploitation
when fewer rounds remain, we refine its samples using the known
synthetic objective. We define
\begin{equation}
    T(h,H)
    =
    T_{\min}+(T_{\max}-T_{\min})\frac{h-1}{H-1},
    \qquad
    K_{h,H}
    =
    \left\lfloor K\frac{T_{\max}}{T(h,H)}\right\rfloor,
    \label{eq:remaining_round_temperature}
\end{equation}
where \(0<T_{\min}\leq T_{\max}\) and \(K\) is the base number
of refinement steps. Starting from
\(\mathbf{x}^{(0)}\sim p_{\mathrm{TS}}(\cdot\mid\mathcal{D})\),
we apply the Langevin-inspired updates
\vspace{-.2cm}
\begin{equation}
    \mathbf{x}^{(k+1)}
    =
    \mathbf{x}^{(k)}
    +\eta\,\mathbf{g}_\tau(\mathbf{x}^{(k)})
    +\sigma_{\mathrm{L}}\sqrt{2\eta}\,\boldsymbol{\xi}_k,
    \qquad
    \boldsymbol{\xi}_k
    \overset{\mathrm{i.i.d.}}{\sim}
    \mathcal{N}(\mathbf{0},\mathbf{I}_{d_\tau}),
    \label{eq:remaining_round_refinement}
\end{equation}
for \(k=0,\ldots,K_{h,H}-1\), where
\(\mathbf{g}_\tau\) is the normalized objective gradient,
\(\eta>0\) is the step size, and \(\sigma_{\mathrm{L}}>0\)
controls the perturbations. Normalization bounds the deterministic step length, while the Gaussian perturbations keep refinement stochastic.
A smaller remaining-round fraction increases the number of updates, rather than their individual magnitude: this allows for objective-guided refinement near the end of optimization.

\textbf{Flow-matching target.}
The finite-step refinement above induces the horizon-dependent distribution
\begin{equation}
    p_1(\cdot \mid \mathcal{D},h,H)
    =
    \mathcal{L}\!\left(\mathbf{x}^{(K_{h,H})}\right).
    \label{eq:refined_target_distribution}
\end{equation}
Although refinement moves TS locations toward higher objective values, the probability mass of \(p_1\) still depends on how frequently the TS initialization and stochastic refinement reach each region.
We therefore define target distribution by favoring high-value regions via the Gibbs variational problem
\begin{equation}
\begin{aligned}
    \tilde p_1
    &=
    \argmax_{p \ll p_1}
    \left\{
        \mathbb{E}_{p}[f_\tau(\mathbf{x})]
        -
        T(h,H)\operatorname{KL}(p\|p_1)
    \right\}
    \propto
    p_1(\cdot \mid \mathcal{D},h,H)
    \exp\!\left(\frac{f_\tau(\cdot)}{T(h,H)}\right).
\end{aligned}
\label{eq:tilted_refined_target_distribution}
\end{equation}
As \(h\) decreases, the same schedule both increases the refinement budget \(K_{h,H}\) and lowers the temperature, progressively shifting the target toward high-value regions.
In practice, we approximate \(\tilde p_1\) by value-weighting samples from \(p_1\), yielding the weighted flow-matching objective in Section~\ref{sec:model}. The variational derivation and sampling approximation are given in Appendices~\ref{app:value_weighting} and~\ref{app:weighted_training}.
The synthetic objective and its gradients are used only during pretraining: deployment requires only the observed history and remaining-round fraction.

\subsection{Model architecture and training objective}
\label{sec:model}

We learn a proposal
\(p_{\theta}(\mathbf{x}\mid\mathcal{D},\alpha)\)
from the target samples constructed in
Section~\ref{sec:target_distribution}, where
\(\alpha=h/H\) is the fraction of optimization rounds remaining.
The model combines a transformer encoder for the observed history with
a conditional flow decoder operating on continuous query locations.

\textbf{History-conditioned architecture.}
Each observation \((\mathbf{x}_i,y_i)\) is embedded using its location
and objective value, then processed by a transformer encoder without
positional encodings~\citep{vaswani2017attention}. The encoded observations
form a memory matrix
\(\mathcal{M}\in\mathbb{R}^{n_c\times d_{\mathrm{model}}}\), where
\(n_c=|\mathcal{D}|\) and \(d_{\mathrm{model}}\) is the hidden dimension.
Each flow state \(\mathbf{x}_t\), at flow time \(t\in[0,1]\), is embedded
and attends to this memory through a cross-attention decoder. Flow time
and the remaining-round fraction modulate the decoder through adaptive
layer normalization, following the DiT architecture~\citep{peebles2023scalable}.
We omit self-attention between query states, allowing independent processing given the history.
For \(M\) flow states, this gives a cross-attention cost of
\(\mathcal{O}(M n_c)\) per decoder evaluation.
The memory is computed once per
optimization round and reused throughout generation. 
Zero-padding and an
active-dimension mask accommodate objectives of dimension \(d\leq D\),
where \(D\) is the maximum supported dimension.
Appendix~\ref{app:model_architecture} specifies the embeddings and decoder
blocks.

\textbf{Velocity distribution.}
Under the straight-line interpolation used for flow matching (Equation~\ref{eq:CFM_loss}), 
the same intermediate state can arise from different source-target pairs and therefore be associated with different target velocities. We model this conditional variation with a Gaussian mixture velocity head.
Following~\citet{chen2025gaussian}:
\vspace{-.3cm}
\begin{equation}
    v_{\theta}
    \left(
        \mathbf{u}\mid\mathbf{x}_t,t,\mathcal{D},\alpha
    \right)
    =
    \sum_{c=1}^{C}
    \pi_c
    \mathcal{N}
    \left(
        \mathbf{u};
        \boldsymbol{\mu}_c,
        \operatorname{diag}(\boldsymbol{\sigma}_c^2)
    \right),
    \label{eq:gmm}
\end{equation}
where \(C\) is the number of components. The mixture weights
\(\pi_c\), means \(\boldsymbol{\mu}_c\), and diagonal standard
deviations \(\boldsymbol{\sigma}_c\) are predicted from the decoded
flow token. Here, \(v_\theta\) denotes a density over velocities,
distinct from the proposal \(p_\theta\) over query locations.

\textbf{Training objective.}
For each sampled episode and history, the ideal objective is the flow-matching
loss under the tilted target \(\tilde p_1\),
\begin{equation}
    \mathcal{L}_{\mathrm{tilt}}(\theta)
    =
    -\mathbb{E}
    \left[
        \mathbb{E}_{\substack{
            t\sim\mathcal{U}(0,1),\,
            \tilde{\mathbf{x}}_1\sim\tilde p_1,
            \mathbf{x}_0\sim p_0
        }}
        \left[
            \log
            v_\theta
            \left(
                \tilde{\mathbf{u}}_t
                \mid
                \tilde{\mathbf{x}}_t,t,\mathcal{D},\alpha
            \right)
        \right]
    \right],
    \label{eq:new_training_objective}
\end{equation}
where \(p_0=\mathcal{N}(\mathbf{0},\mathbf{I}_d)\),
\(\tilde{\mathbf{x}}_t=(1-t)\mathbf{x}_0+t\tilde{\mathbf{x}}_1\), and
\(\tilde{\mathbf{u}}_t=\tilde{\mathbf{x}}_1-\mathbf{x}_0\).
The outer expectation is over sampled episodes and histories.
Rather than sampling directly from \(\tilde p_1\), we approximate the inner
expectation using samples
\(\mathbf{x}_1^{(i)}\sim p_1(\cdot\mid\mathcal{D},h,H)\)
and self-normalized importance sampling
\citep{aeckerle2026error}. With
\(\mathbf{x}_t^{(i)}=(1-t)\mathbf{x}_0^{(i)}+t\mathbf{x}_1^{(i)}\) and
\(\mathbf{u}_t^{(i)}=\mathbf{x}_1^{(i)}-\mathbf{x}_0^{(i)}\), this gives
\begin{equation}
    \mathcal{L}(\theta)=
    -\mathbb{E}
    \left[
        \sum_{i=1}^{M}
        w_i
        \log
        v_{\theta}
        \left(
            \mathbf{u}_t^{(i)}
            \mid
            \mathbf{x}_t^{(i)},t,\mathcal{D},\alpha
        \right)
    \right], \quad w_i=\operatorname{softmax}_i
    \left(f_\tau(\mathbf{x}_1^{(i)})/T(h,H)\right).
    \label{eq:training_objective}
\end{equation}
Appendix~\ref{app:weighted_training} derives this approximation, while lastly,
Section~\ref{sec:closed_loop_training} describes history collection.

\textbf{Batch generation.}
At deployment, we transform \(q\) independent Gaussian base samples
through the learned flow, sharing the encoded history across their
trajectories. 
The model parameters remain fixed, and the requested
batch size determines the number of samples generated. See Algorithm \ref{alg:fmtr_inference} for batch generation procedure. 

\vspace{-.2cm}
\subsection{Closed-loop pretraining}
\label{sec:closed_loop_training}

The histories encountered at deployment depend on batches generated by
FMTR in previous rounds. Training only on randomly sampled histories or
trajectories generated solely from \(\tilde{p}_1\) would therefore expose the model
to a different state distribution. We address this with Dataset Aggregation (DAgger,~\citealt{ross2011reduction}). At rollout iteration \(s\), the
entire batch at each visited state is sampled either from the target
distribution or from the current FMTR proposal:
\begin{equation}
    B_r
    \sim
    \begin{cases}
        \tilde{p}_1\!\left(
            \cdot\mid\mathcal{D}_r,h_r,H
        \right)^{\otimes q},
        & \text{with probability } \beta_s, \\[2mm]
        p_{\theta}\!\left(
            \cdot\mid\mathcal{D}_r,\alpha_r
        \right)^{\otimes q},
        & \text{with probability } 1-\beta_s,
    \end{cases}
    \qquad
    \alpha_r=\frac{h_r}{H},
    \label{eq:rollout_mixture}
\end{equation}
where \(\beta_s := \gamma^s, \gamma \in (0,1) \), decreases over rollout iterations.
Regardless of which
distribution advances the trajectory, \(\tilde{p}_1\) is recomputed at every
visited state using the corresponding synthetic episode and stored as
supervision. The collected histories therefore become increasingly
representative of those induced by FMTR while retaining target samples at
each visited state. 
In practice, for simplicity, we can use $p_1$ instead of $\tilde{p}_1$ for $B_r$.

Moreover, since episodes with different values of \(n_0\), \(q\), and \(H\) produce context
sizes
\(\lvert\mathcal{D}_r\rvert=n_0+rq\)
at unequal frequencies, we  partition the replay buffer into
buckets indexed by context size. Each training minibatch is obtained by
first sampling a non-empty bucket uniformly and then sampling examples
from that bucket, preventing frequently occurring context lengths from
dominating training. Parameter updates use the current replay buffer,
which is periodically augmented with rollouts from the updated model. The
exact
complete training
procedure is given in Appendix~\ref{app:closed_loop_training}.
Algorithm~\ref{alg:fmtr_training} summarizes the collection and training
cycle, see Algorithm~\ref{alg:rollout_collection} for the full procedure.
Figure~\ref{fig:full_process} 
shows the targets and the learned proposal 
over a full optimization trajectory.

\begin{algorithm}[h!]
\caption{FMTR pretraining: overview}
\label{alg:fmtr_training}
\small
\begin{algorithmic}[1]
\Require Episode prior $p(\tau)$; update budget $N_{\mathrm{train}}$;
rollout settings $N_{\mathrm{roll}},M,T_{\mathrm{refresh}},\gamma$

\State Initialize $\theta$, bucketed buffer $\mathcal{B}\leftarrow\varnothing$,
and counters $u\leftarrow0$, $s\leftarrow0$

\While{$u<N_{\mathrm{train}}$}
    \State $\beta_s\leftarrow\gamma^s$
    \Comment{$\beta_0=1$: initial rollouts use $p_1$ only}

    \For{each of $N_{\mathrm{roll}}$ episodes $\tau\sim p(\tau)$}
        \State $\mathcal{D}\leftarrow\mathcal{D}_0^\tau$

        \For{$r=0,\ldots,H_\tau-1$}
            \State $h\leftarrow H_\tau-r$, $\alpha\leftarrow h/H_\tau$

            \State Draw $M$ Thompson-sampling locations
            (\Eqref{eq:task_ts_distribution})

            \State Refine the Thompson-sampling locations for
            $K_{h,H_\tau}$ steps (\Eqref{eq:remaining_round_refinement})

            \State Store the state, refined targets, and their values
            in $\mathcal{B}_{|\mathcal{D}|}$

            \State Sample $B_r$ from the target--model mixture
            (\Eqref{eq:rollout_mixture})

            \State Evaluate $B_r$ in parallel and append its
            observations to $\mathcal{D}$
        \EndFor
    \EndFor

    \State $U\leftarrow\min(T_{\mathrm{refresh}},N_{\mathrm{train}}-u)$

    \State Perform $U$ updates using context-length-stratified replay
    (\Eqref{eq:training_objective})

    \State $u\leftarrow u+U$, $s\leftarrow s+1$
\EndWhile

\State \Return $\theta$
\end{algorithmic}
\end{algorithm}
\vspace{-.2cm}
\section{Related Work}
\vspace{-.2cm}

\textbf{Batch Bayesian optimization.}
Classical batch BO either optimizes set-valued acquisitions, Monte Carlo batch acquisitions, and energy--entropy criteria, or constructs batches through sequential or independent single-point decisions~\citep{wilson2018maximizing,balandat2020botorch,teufel2024,gonzalez2016batch,kandasamyparallelts}. Joint acquisitions account for interactions among $q$ candidates but require optimization over $qd$ variables; local penalization and parallel TS replace this with repeated or independent optimizations. These methods retain a task-specific surrogate and online design-space search at each round.

\textbf{Generative models for BO.}
Recent work uses generative models directly as proposal distributions for batch BO. DiBO learns a diffusion prior over evaluated designs and an ensemble surrogate for uncertainty-aware utility, then fine-tunes the sampler toward their product distribution before applying local search and filtering~\citep{yun2025posterior}. GenBO removes the intermediate surrogate by fitting a single generative model directly from noisy utilities, targeting a proposal whose density follows an expected-utility acquisition~\citep{oliveira2026generative}. Both update task-specific generators as observations accumulate, whereas FMTR pretrains a history-conditioned proposal that adapts in context with fixed parameters.

\textbf{In-context optimization.}
In-context methods replace different components of the BO loop with pretrained models. PFNs4BO and GIT-BO perform posterior prediction in context, while $\alpha$-PFN predicts entropy-search values~\citep{muller2023pfns4bo,yu2026git,rakotoarison2026}. These methods retain an explicit acquisition or candidate-search step. End-to-end methods instead learn the decision rule from optimization trajectories, including NAP, PABBO, TAMO, and FICBO~\citep{maraval2023endtoend,zhang2025pabbo,zhang2026incontext,blumer2026ficbo}. These policies generally act on supplied candidate sets, but FMTR  learns a continuous history-conditioned sampler for variable-size batches.


\vspace{-.3cm}
\section{Experiments}
\label{sec:experiments}
\vspace{-.3cm}
\begin{figure}[h!] 
    \centering
    \includegraphics[width=\linewidth]{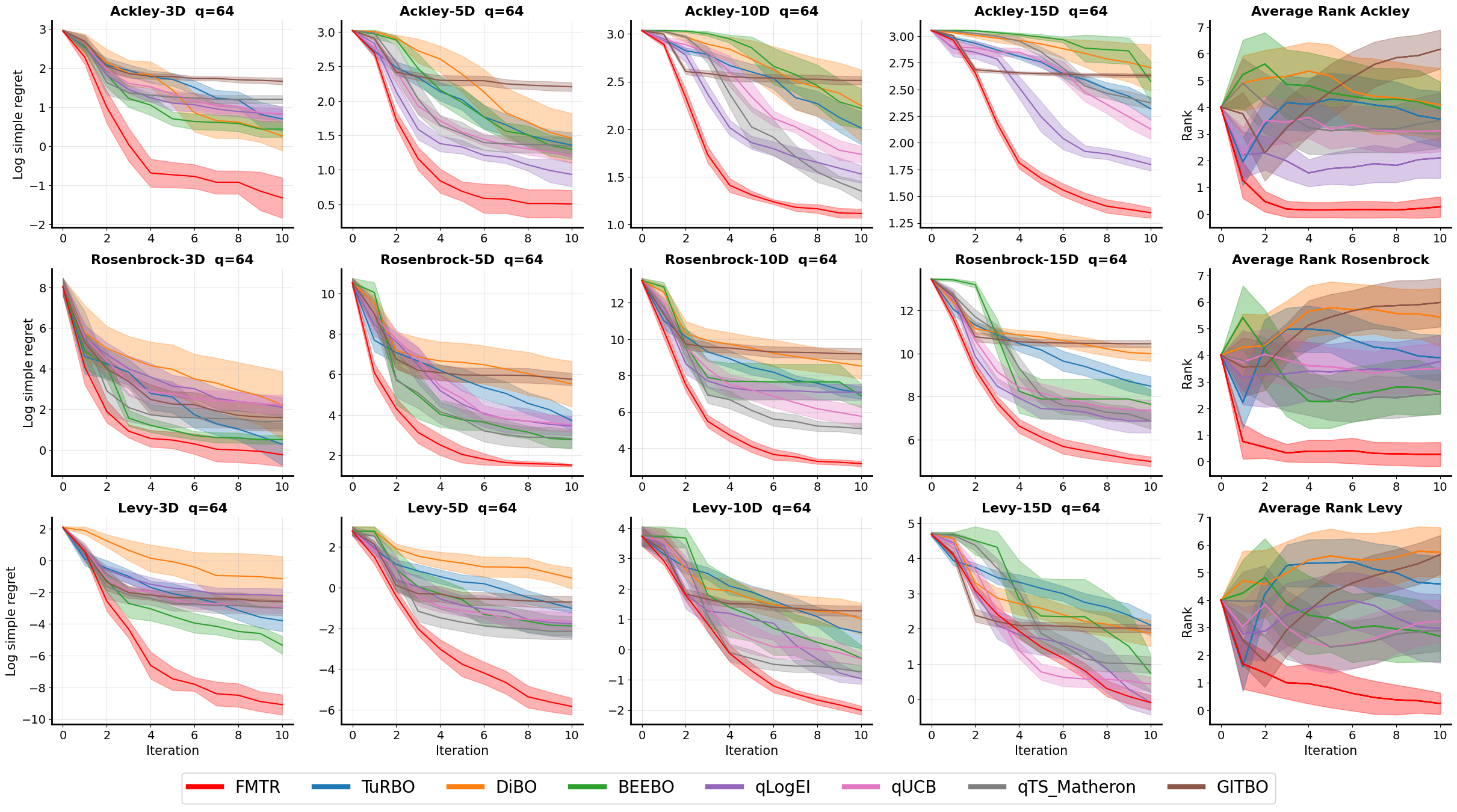}
\caption{\textbf{Experiments on synthetic functions.} Log simple regret over 15 runs, shown as mean with 95\% CI. The last column reports the average rank across dimensions for each function.}
    \label{fig:synthetic}
\end{figure}
\vspace{-.3cm}
\textbf{Baselines.}
We compare FMTR with the trust-region optimizer
TuRBO~\citep{eriksson2019scalable},
BEEBO~\citep{teufel2024}, the diffusion-based optimizer
DiBO~\citep{yun2025posterior}, and the in-context surrogate
method GIT-BO~\citep{yu2026git}.
We also include three GP-based batch strategies:
qLogEI~\citep{ament2023unexpected}, parallel TS with Matheron's rule
(qTS-Matheron;~\citealp{kandasamyparallelts,wilson2020efficiently}),
and qUCB~\citep{wilson2018maximizing}.
See Appendix~\ref{app:baselines} for baseline details.

\textbf{Implementation details and evaluation protocol.}
Each run starts from \(10\) observations and proceeds for
\(H=10\) batch-evaluation rounds, with batch sizes specified
below. We apply the Yeo--Johnson transformation to observed
objective values~\citep{cowen2012hebo}.
FMTR's parameters remain fixed during evaluation, and each
batch is generated using \(100\) flow-integration steps without
access to test-objective gradients.
We report means and \(95\%\) confidence intervals over
\(15\) independent runs.
Pretraining settings and model details are provided in
Appendices~\ref{app:pretraining_details}
and~\ref{app:model_details}.

\vspace{-.2cm}
\subsection{Results}
\vspace{-.1cm}

\begin{figure}[h] 
    \centering
    \includegraphics[width=\linewidth]{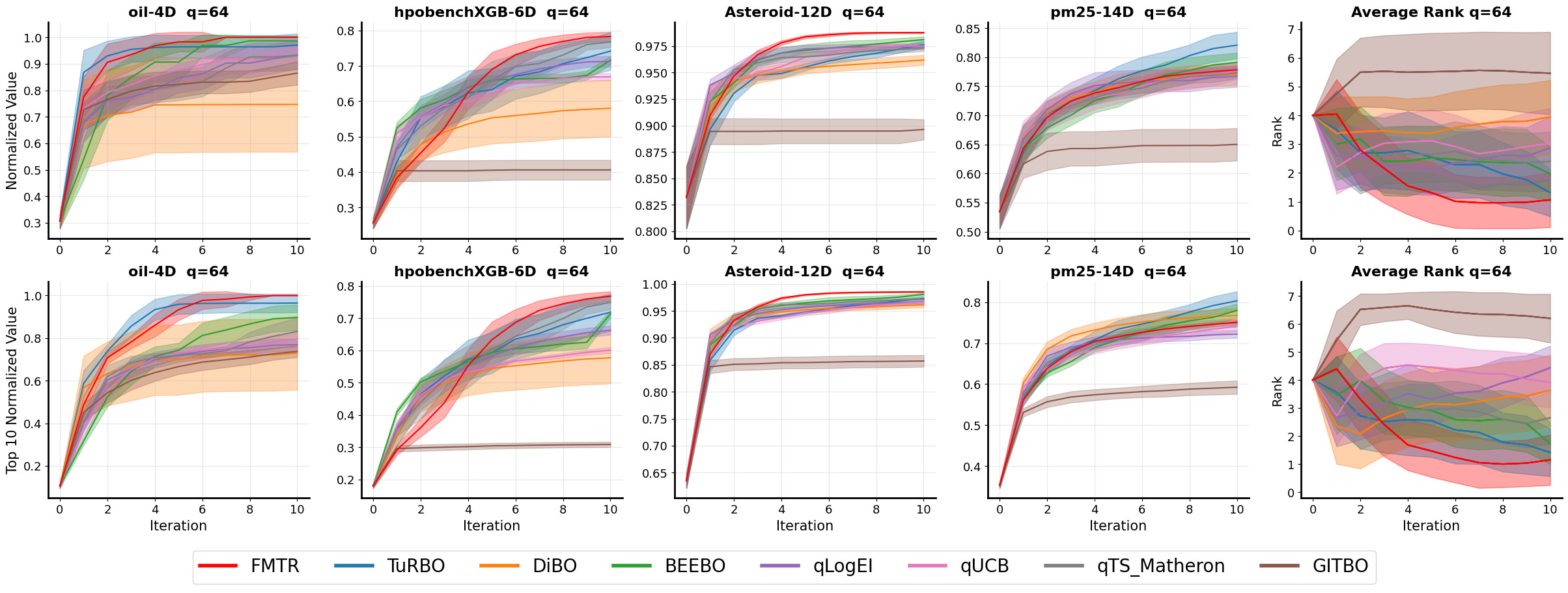}
    \caption{\textbf{Experiments on real-world functions}. \textbf{Top row:} normalized best value. \textbf{Bottom row:} Top 10 normalized best value. Mean with 95\% CI computed across 15 runs with random initializations. Higher is better. Last column is the average rank across all tasks for each batch size.}
    \label{fig:realworld}
\end{figure}

\begin{figure}[h]
    \centering
    \includegraphics[width=\linewidth]{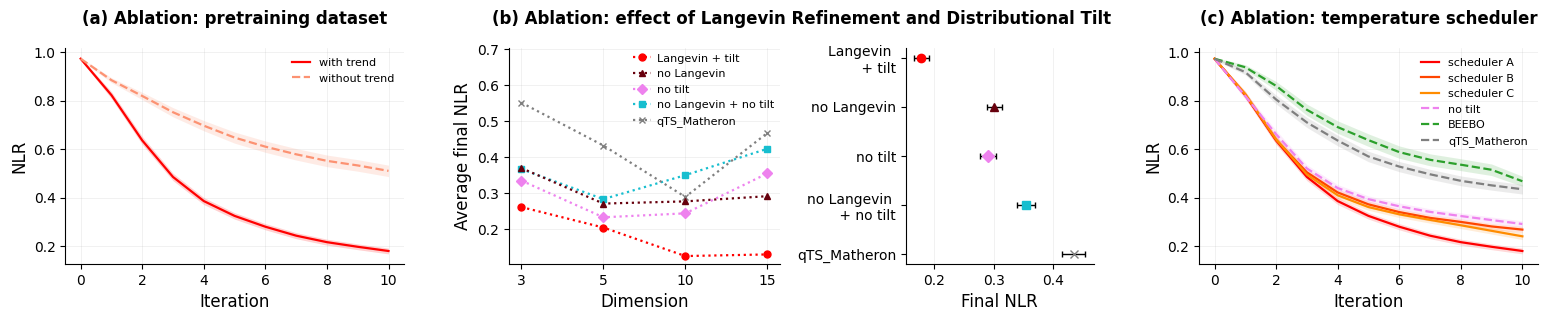}
    \caption{\textbf{Ablation studies.}
Performance is normalized and averaged on the Ackley, Levy, and Rosenbrock functions with $q \in \{32,64\}$ and $d \in \{3,5,10,15\}$. 
\textbf{(a)} Effect of the pretraining dataset composition. 
\textbf{(b)} Effect of the Langevin refinement and Distributional Tilt.
\textbf{(c)} Effect of temperature scheduler. These results represent averages, individual results can be found in Figure~\ref{fig:detail_ablation}.}
    \label{fig:new_ablation}
\end{figure} 

\textbf{Synthetic functions.} We evaluate Ackley, Rosenbrock, and Levy with
\(d\in\{3,5,10,15\}\) and batch size \(q=64\) 
(Figure~\ref{fig:synthetic}). 
FMTR achieves the best final average rank for each function,
consistently across tested dimensions.
Although geared towards \(q=1\),
GIT-BO supports batch selection through multiple effective subspaces, but its performance degrades markedly in the large-batch regime considered here. More explicitly batch-oriented alternatives also remain behind FMTR.
BEEBO directly constructs joint batches through an energy--entropy acquisition,
explicitly accounting for dependencies among candidates, 
while DiBO use a
diffusion-based generative model to form batch proposals. In DiBO, 

accurately
learning 
the acquisition from the limited observations available
during optimization can be difficult, potentially hindering its ability to identify
promising regions. FMTR instead grounds its target distribution in TS and amortizes the subsequent refinement through offline pretraining,
combining posterior-driven exploration with knowledge transferred across
optimization tasks. Moreover, FMTR consistently outperforms qTS, even though TS provides the base distribution from which its training targets are constructed.
We return to this finding in Section~\ref{sec:ab}, where we examine which components of FMTR account for this improvement. Extended results for \(q\in\{16,32\}\) and more synthetic functions show broadly similar trends, although FMTR's advantage is more pronounced for larger batches (Appendix~\ref{app:additional_experiments}). Finally, FMTR achieves the lowest reported runtime among the methods (Table~\ref{tab:runtime_comparison}).

\textbf{Real-world applications.}
We evaluate four data-derived benchmarks considered by
\citet{hsieh2021reinforced}: oil-4D,
hpobenchXGB-6D, Asteroid-12D, and
pm25-14D (Appendix~\ref{app:realworld_tasks}) with batch sizes \(q=64\).
Figure~\ref{fig:realworld} reports the best normalized objective value and
the Top 10 Normalized Value, i.e., the mean normalized value of the ten
best points observed so far. Values are scaled by
\((f(\mathbf{x})-f_{\min})/(f_{\max}-f_{\min})\).
The latter metric measures the ability to identify multiple high-value candidates.
FMTR remains competitive across these tasks and often
attains the highest values in later rounds. Compared with the
synthetic benchmarks, however, its advantage is less uniform,
with several baselines closely matching its performance, such as 
TuRBO, which even obtains higher mean
values on PM2.5.
Nevertheless, FMTR achieves the best final
average rank under both metrics. Extended results for $q=32$ in Appendix~\ref{app:real_world} show a similar trend, with FMTR achieving stronger performance on both metrics in general. These results support the transfer of the
pretrained proposal to real-world objectives, albeit
with task-dependent differences in performance.

\vspace{-.1cm}
\subsection{Ablation Studies}
\label{sec:ab}
We examine whether different modeling choices and hyperparameter settings significantly affect the performance of FMTR. In all experiments in this section, we evaluate FMTR and all baselines on the Ackley, Levy, and Rosenbrock functions with $q \in \{32,64\}$, $d \in \{3,5,10,15\}$, and $H=10$. To average performance across functions with different regret scales, we report Normalized Log Simple Regret (NLR). Further details are provided in Appendix~\ref{app:ablation_studies}.

\textbf{Pretraining dataset composition.}
We compare FMTR pretrained with the full synthetic objective in Equation~\ref{eq:synthetic_objective} against a variant using only the RFF component. Figure~\ref{fig:new_ablation}(a) reports mean with 95\% CI across the three synthetic functions. Including the trend component substantially
improves performance. One explanation is that broadly sampled GP hyperparameters can otherwise
produce highly irregular functions, for example Mat\'ern-\(3/2\) samples with short lengthscales, that are less representative of practical optimization landscapes. The random trend adds smoother global structure while preserving the local variability of the RFF component.

\textbf{Effect of Langevin refinement and distributional tilt.}
We disentangle the two transformations of the TS target by comparing FMTR with three variants:
\emph{no Langevin}, which applies the exponential tilt directly to the unrefined TS target;
\emph{no tilt}, which retains Langevin refinement but trains directly on \(p_1\);
and \emph{no Langevin + no tilt}, which amortizes the unrefined TS target.
Figure~\ref{fig:new_ablation}(b) shows that removing either refinement or tilt causes a clear, similar degradation, while removing both yields the weakest FMTR variant. This indicates complementary gains from the two components. Still, the doubly ablated model outperforms qTS, pointing to an additional benefit from amortized pretraining.

\textbf{Effect of the temperature schedule.}
We compare three schedules for \(T(h,H)\): \emph{scheduler A}, from
Equation~\ref{eq:remaining_round_temperature};
\emph{scheduler B},
\(
T(h,H)
=
T_{\max}
\left(
T_{\min}/T_{\max}
\right)^{1-\frac{h-1}{H-1}}
\);
and \emph{scheduler C},
\(
T(h,H)
=
\mathbb{I}\{h=1\}(T_{\min}-T_{\max})+T_{\max}
\),
following~\citealt{teufel2024}.
Figure~\ref{fig:new_ablation}(c) shows that all three schedules outperform the variants without distributional tilt, indicating that the gains are not limited to a single temperature schedule.
Still, \emph{scheduler A} achieves the lowest NLR in later rounds, suggesting that the precise annealing profile still affects the strength and timing of exploitation.

\section{Conclusion}

We introduced FMTR, an in-context optimizer for large-batch, few-round
black-box optimization that learns a continuous query proposal from
horizon-dependent refinements of TS locations.
Conditional flow matching amortizes these targets into parallel batch
generation, avoiding task-specific model fitting and acquisition
optimization. Across benchmarks, FMTR combines strong optimization
performance at low online cost, while ablations show that both target
refinement and value weighting improve over parallel TS.
Notably, a single pretrained model remains competitive from \(d=2\) to \(15\), reusing optimization experience across tasks and search-space dimensions, whereas task-specific batch baselines fit each objective independently.

\textbf{Limitations and future work.}
An interesting point is that FMTR samples batch members independently given the history and remaining
budget, without explicitly modeling within-batch redundancy. Although it
remains competitive with joint-batch methods like BEEBO, joint
proposals could further improve complementarity between evaluations. However, this
would require jointly structured pretraining targets and an explicit
batch-level objective, thus increasing pretraining complexity.
Performance also depends on the synthetic-task prior (Section~\ref{sec:ab}), motivating better-designed training
distributions. Extending FMTR beyond \(D=15\) and to discrete domains are
further directions.

\section*{Reproducibility statement}
We provide all experimental details needed for reproducibility. Hyperparameters are specified in Section~\ref{app:kernel} and Table~\ref{tab:pretraining_task_prior}; the pretraining dataset generation procedure is described in Section~\ref{app:pretraining_details}; the complete training algorithm is provided in Algorithm~\ref{alg:rollout_collection}; baseline hyperparameters are reported in Section~\ref{app:baselines}; and the synthetic and real world benchmark functions are described in Sections~\ref{app:synthetic_tasks} and~\ref{app:realworld_tasks}, respectively. The code will be made publicly available upon acceptance.

\section*{AI disclosure}
In this work, we used generative AI tools for providing feedback on research methodology and implementing methods. We have not used generative AI tools for generating synthetic data sets, developing conceptual frameworks, assisting with reformat dataset and interpreting results are not applicable to this work. Additionally, we used generative AI tools for brainstorming, searching for information, and editing a research paper to improve readability. We have reviewed all AI-assisted work. LLM-generated code was verified and tested for correctness by the authors. We take responsibility for the final content of this work, including text, claims or artifacts produced with the aid of generative AI.

\section*{Acknowledgements}
The authors thank Conor Hassan for helpful comments and feedback on early versions of this work. This work was supported by EU funding ERC ODD-ML 101201120 and the Research Council of Finland Flagship programme: Finnish Center for Artificial Intelligence FCAI and decisions 359207, 365982. S. Kaski was additionally supported by the UKRI Turing AI World-Leading Researcher Fellowship (EP/W002973/1) and J. Martinelli was supported by Research Council of Finland (grant agreement 377145). The authors acknowledge the computational
resources provided by the Aalto Science-IT Project from Computer Science IT. Views and opinions expressed are however those of the author(s) only and do not necessarily reflect those of the European Union or the European Research Council Executive Agency. Neither the European Union nor the granting authority can be held responsible for them. 

\vspace{1em} 

\noindent
\begin{minipage}{\textwidth}
  \centering
  \includegraphics[height=1.5cm]{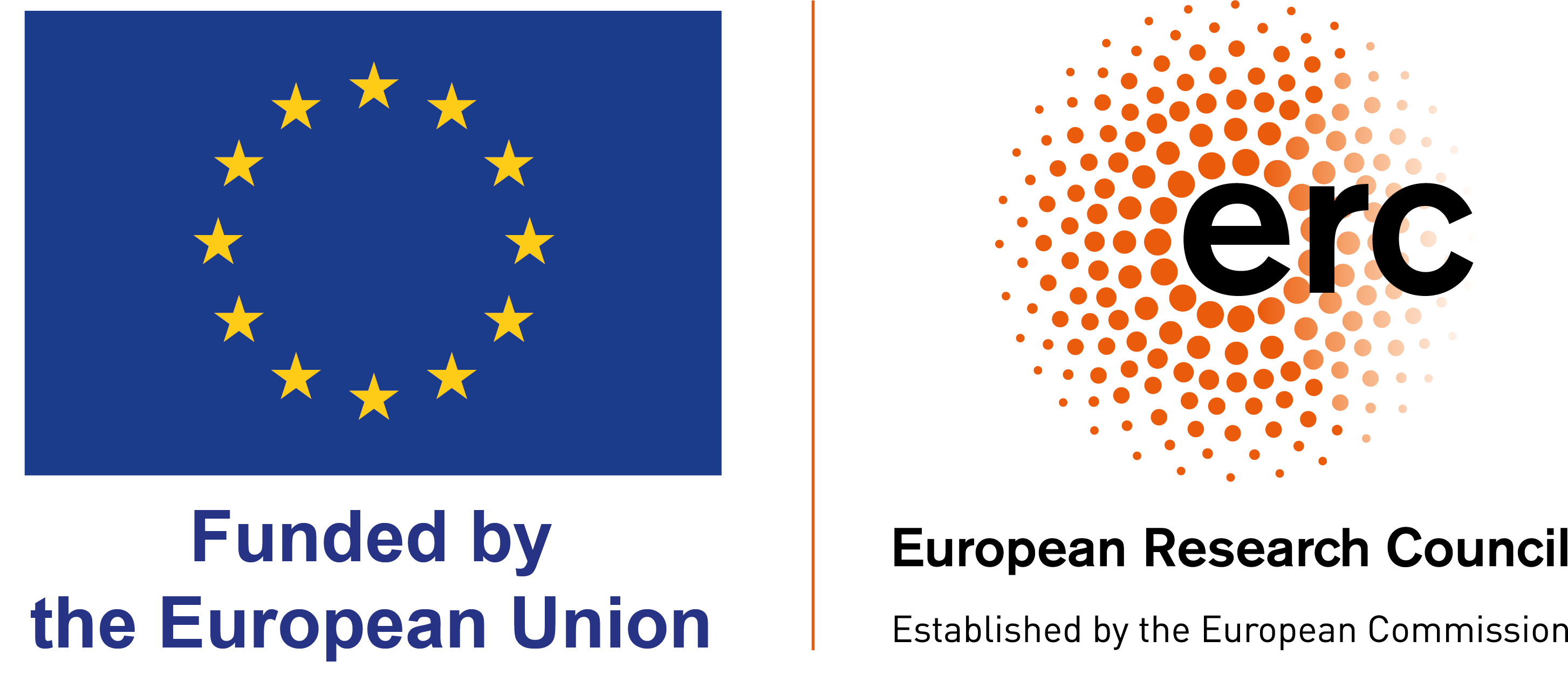}
  \hfill
\end{minipage}

\bibliography{references}
\bibliographystyle{iclr2027_conference}

\clearpage
\newpage
\appendix

\setcounter{figure}{0}
\setcounter{table}{0}
\setcounter{equation}{0} 
\renewcommand{\thefigure}{S\arabic{figure}}
\renewcommand{\thetable}{S\arabic{table}}
\renewcommand{\theequation}{S\arabic{equation}}
\renewcommand{\theHequation}{S\arabic{equation}}

The appendix is organized as follows:
\begin{itemize}
    \item Appendix~\ref{app:additional_experiments} provides additional
    synthetic and real-world results across batch sizes, detailed ablations,
    a visualization of the optimization process, examples of pretraining
    tasks, and runtime comparisons.

    \item Appendix~\ref{app:pretraining_details} details the synthetic task
    prior and closed-loop pretraining procedure.

    \item Appendix~\ref{app:target_construction} gives the pathwise GP posterior
    construction, stochastic refinement, and Gibbs formulation of the
    value-weighted query target.

    \item Appendix~\ref{app:model_details} specifies the model architecture,
    derives the self-normalized importance sampling approximation to the
    tilted-target training objective, and provides batch-generation and
    computational details.

    \item Appendix~\ref{app:experimental_details} provides baseline details
    and definitions of the synthetic and real-world benchmark tasks.
\end{itemize}

\section{Additional Experiments}
\label{app:additional_experiments}

\subsection{More experiments on synthetic functions}

\begin{figure}[H] 
    \centering
    \includegraphics[width=\linewidth]{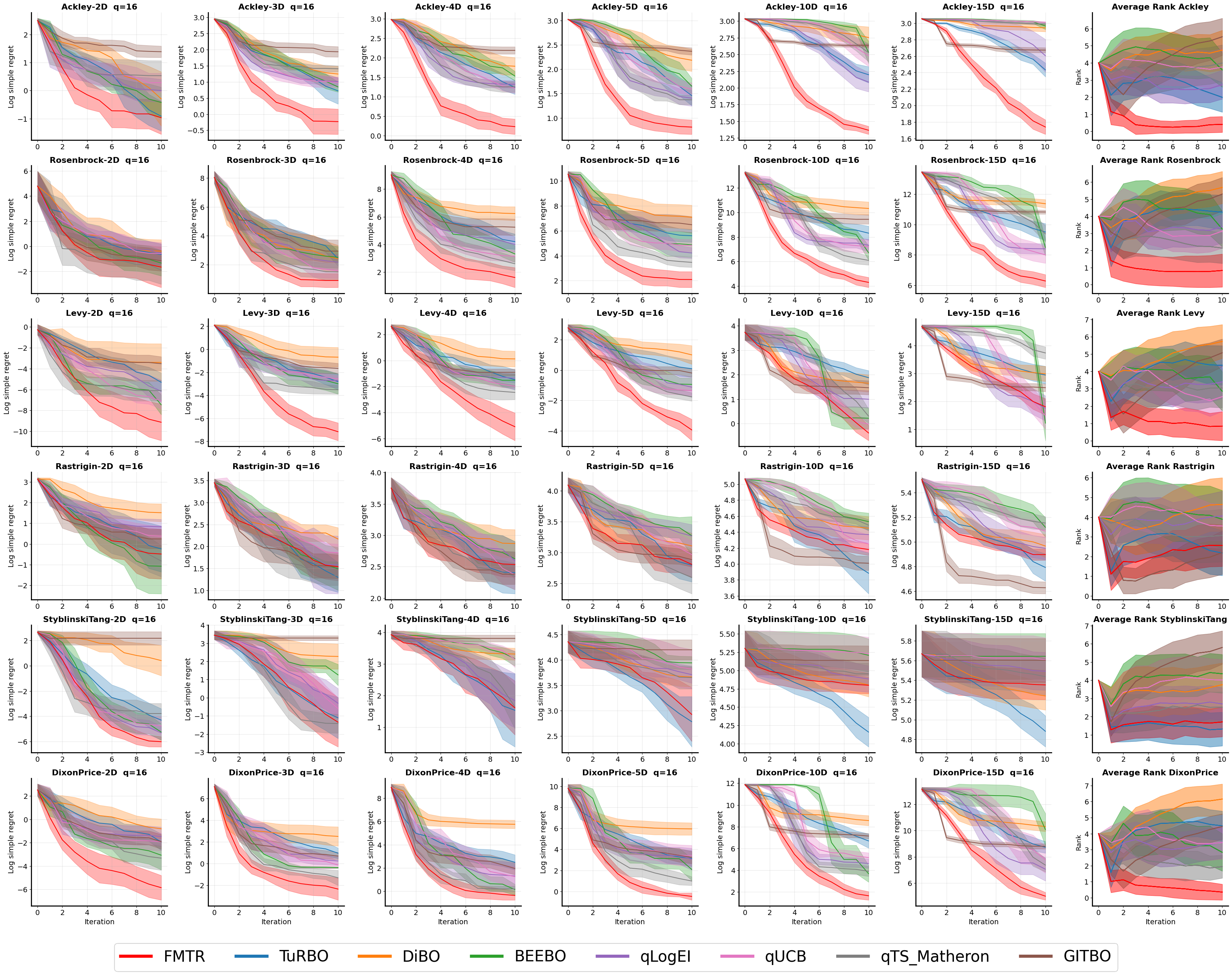}
    \caption{\textbf{Experiments on synthetic functions with $q=16$.} Log simple regret over 15 runs, shown as mean with 95\% CI. The last column reports the average rank across dimensions for each function.}
    \label{fig:full_synthetic_16}
\end{figure}

\begin{figure}[H] 
    \centering
    \includegraphics[width=.9\linewidth]{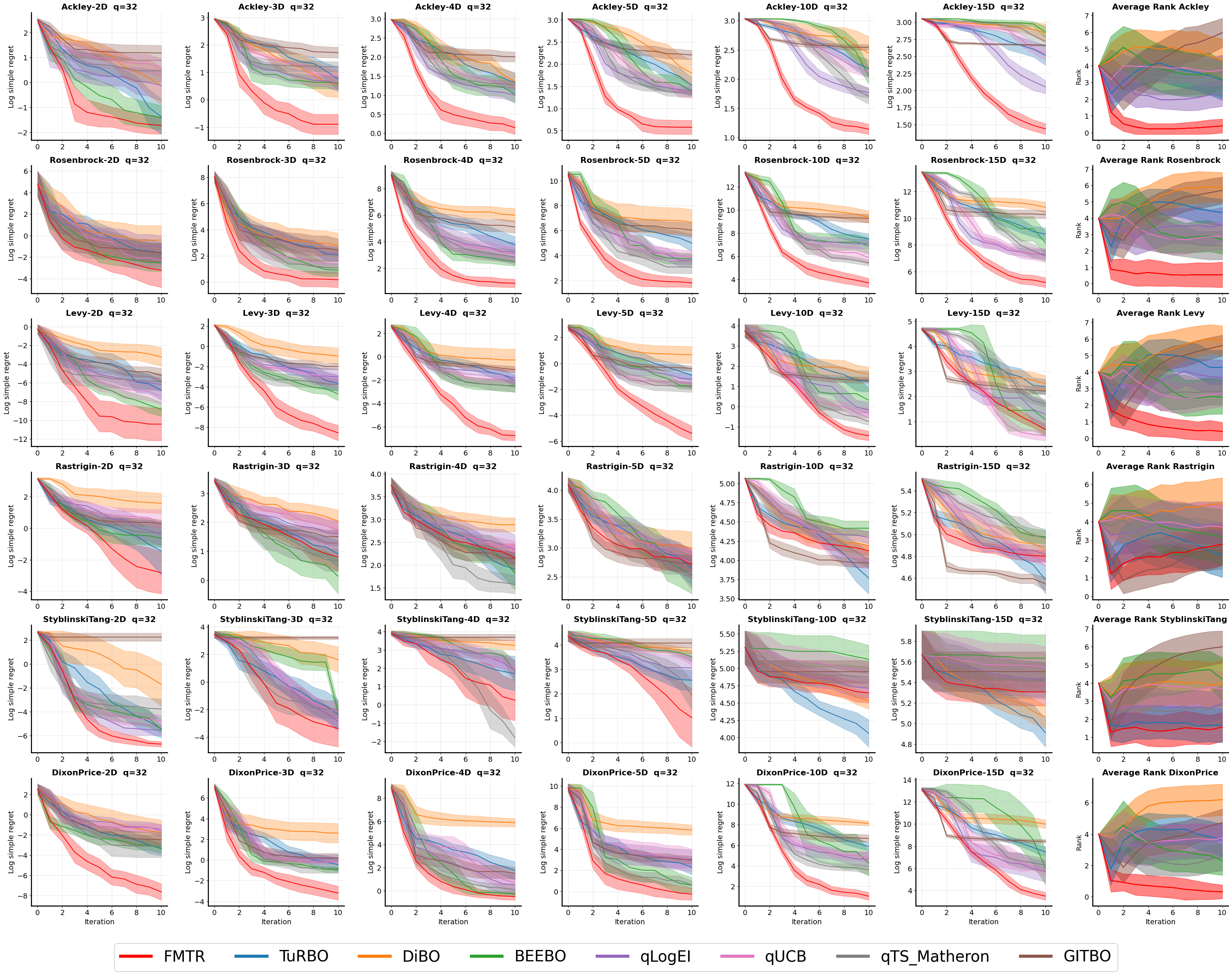}
    \captionsetup{skip=1.6pt}
    \caption{\textbf{Experiments on synthetic functions with $q=32$.} Log simple regret over 15 runs, shown as mean with 95\% CI. Last column reports average rank across dimensions for each function.}
    \label{fig:full_synthetic_32}
\end{figure}

\begin{figure}[H] 
    \centering
    \includegraphics[width=.9\linewidth]{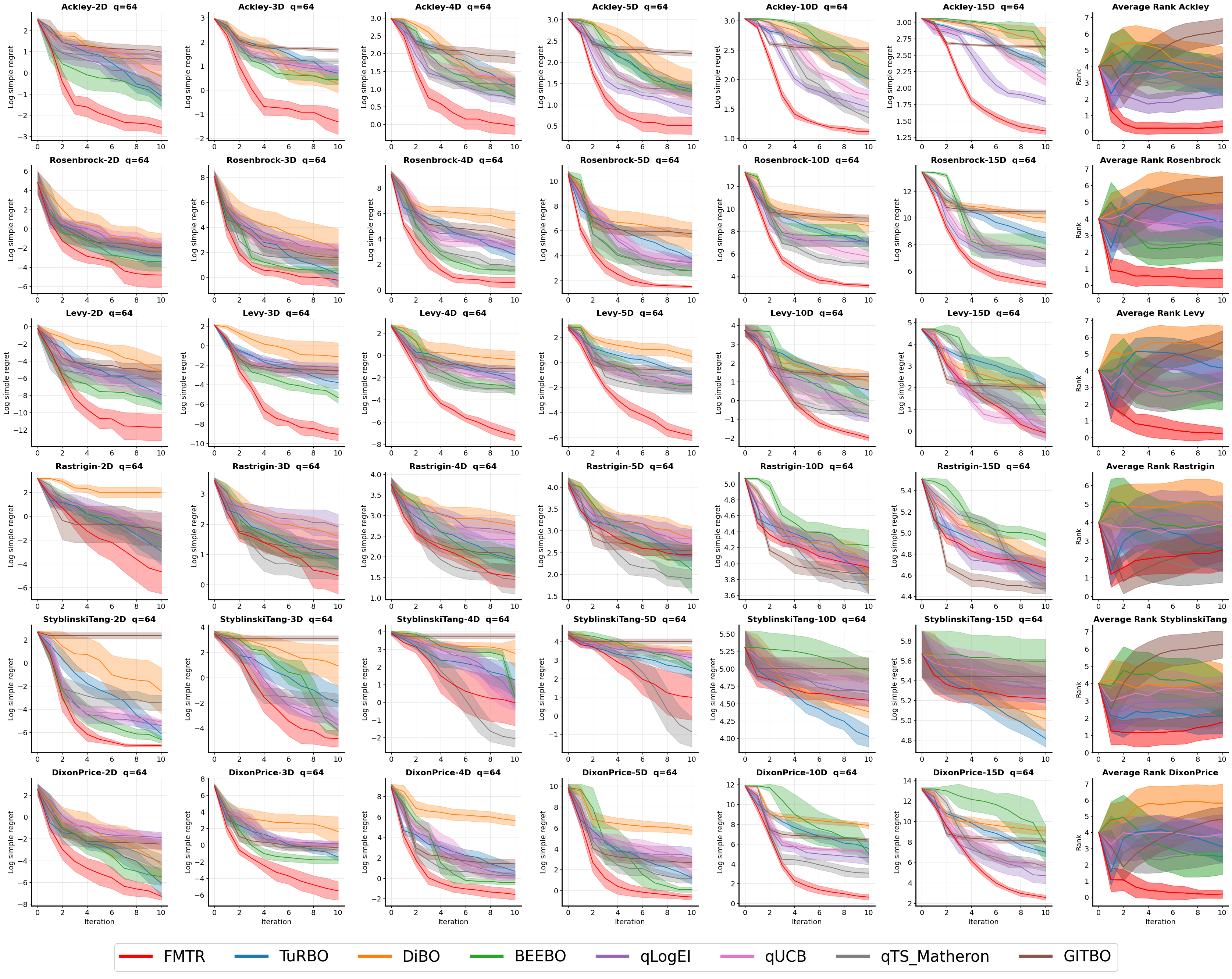}
    \captionsetup{skip=1.6pt}
    \caption{\textbf{Experiments on synthetic functions with $q=64$.} Log simple regret over 15 runs, shown as mean with 95\% CI. Last column reports average rank across dimensions for each function.}
    \label{fig:full_synthetic_64}
\end{figure}

\subsection{Further detail on Ablation Studies} 
\label{app:ablation_studies}
\begin{figure}[H] 
    \centering
    \includegraphics[width=.94\linewidth]{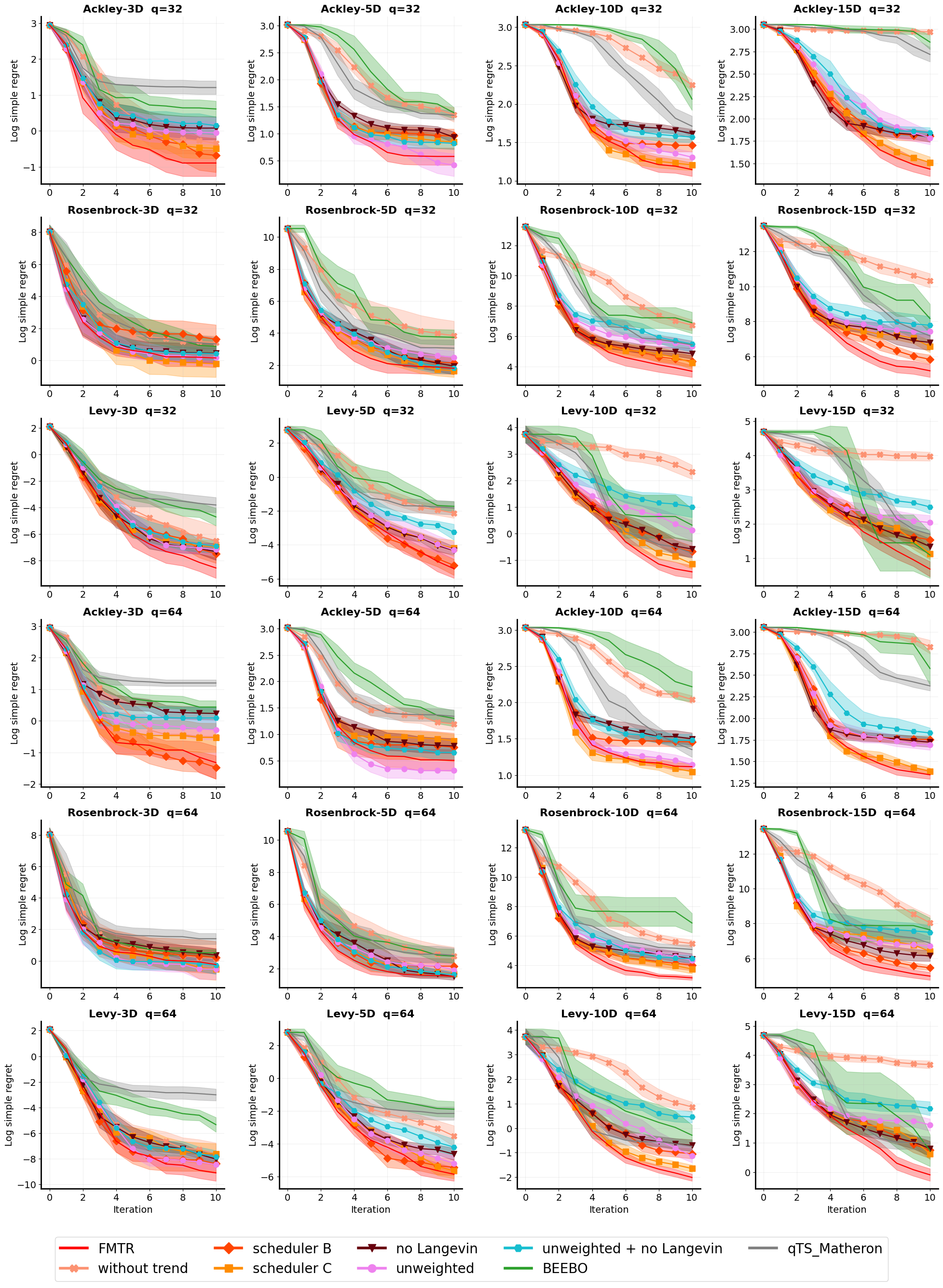}
    \captionsetup{skip=1.8pt}
    \caption{\textbf{More detail on Ablation Studies}. \textbf{Rows 1--3:} Log simple regret with batch size 32. \textbf{Rows 4--6:} Log simple regret with batch size 64. Mean with 95\% CI computed across 15 runs with random initializations. Lower is better.}
    \label{fig:detail_ablation}
\end{figure}

Because the Ackley, Levy, and Rosenbrock functions have different regret
scales, we normalize the log simple regret before averaging performance
across functions in Section~\ref{sec:ab}. We refer to the resulting metric
as Normalized Log Simple Regret (NLR). For each function and dimensionality,
we compute the minimum and maximum log simple regret across the methods
evaluated in Section~\ref{sec:ab} and rescale to \([0,1]:
\mathrm{NLR}
=
\frac{
\text{log simple regret}-\text{min}
}{
\text{max}-\text{min}
}.
\)
The individual results are shown in Figure~\ref{fig:detail_ablation}.

\subsection{Visualization on process}
\begin{figure}[H] 
    \centering
    \includegraphics[width=.92\linewidth]{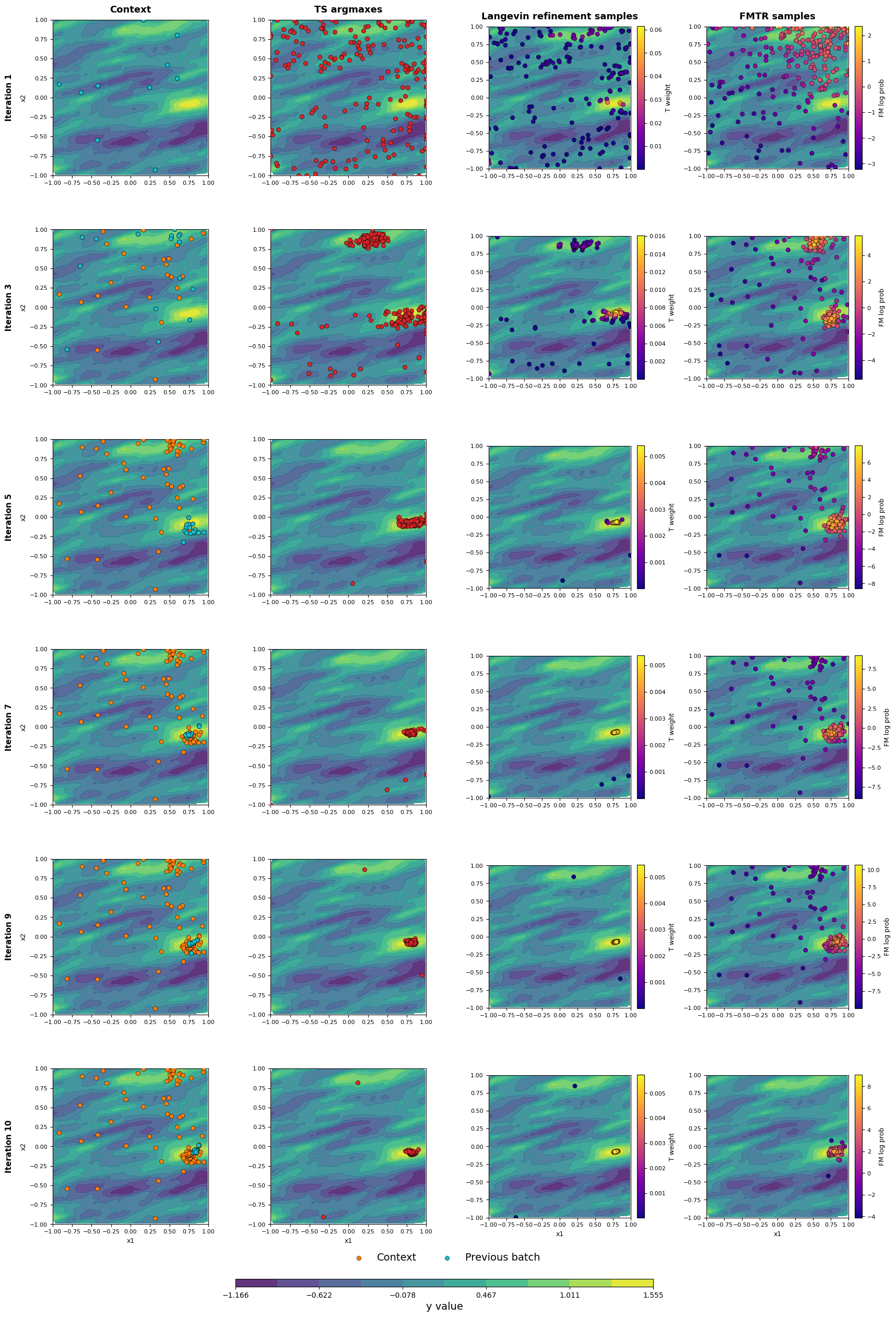} 
    \captionsetup{skip=1.8pt}
\caption{\textbf{Evolution of FMTR over a complete optimization rollout.}
Example two-dimensional trajectory with \(q=16\) and \(H=10\), shown at iterations \(1,3,5,7,9,\) and \(10\).
From left to right: the accumulated context with the previous batch highlighted, TS argmaxes conditioned on the current history, their horizon-dependent Langevin refinements, and samples from the learned FMTR proposal.
The background shows the objective. As observations accumulate and the horizon shrinks, the target and FMTR proposal concentrate on high-value regions.}
    \label{fig:full_process}
\end{figure}

Figure~\ref{fig:full_process} illustrates the targets and the FMTR proposal along one rollout
with $q=16$ and $H=10$.

\subsection{More experiments on real-world applications} \label{app:real_world}

\begin{figure}[H] 
    \centering
    \includegraphics[width=\linewidth]
    {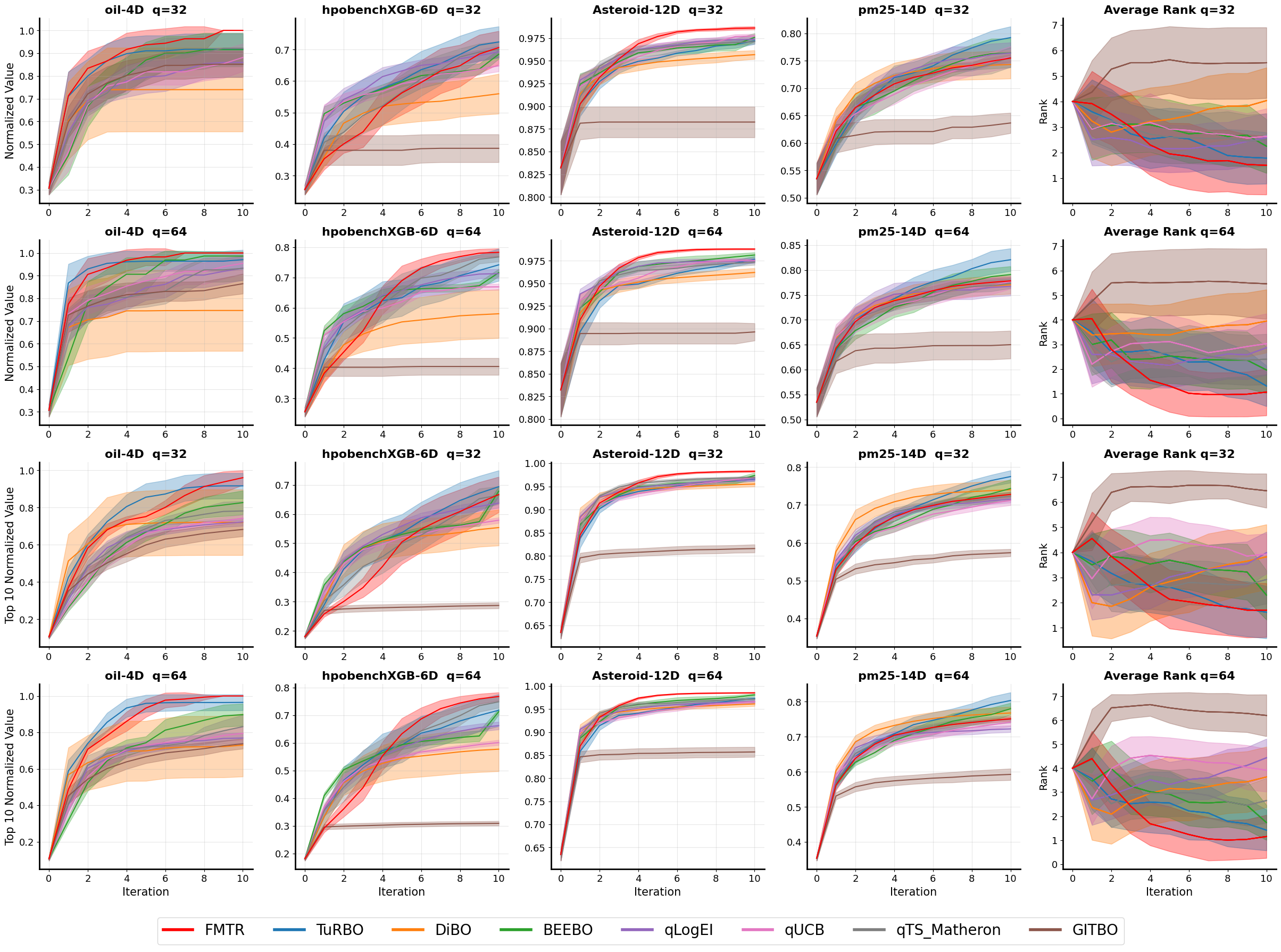}
    \captionsetup{skip=1.8pt}
    \caption{\textbf{Experiments on real-world functions with $q\in \{32,64\}$}. \textbf{Top 2 rows:} normalized best value. \textbf{Bottom 2 rows:} Top 10 normalized best value. Mean with 95\% CI computed across 15 runs with random initializations. Higher is better. Last column is the average rank across all tasks for each batch size.}
    \label{fig:full_realworld}
\end{figure}

\subsection{Examples of training tasks}
\begin{figure}[H] 
    \centering   \includegraphics[width=0.8\linewidth]{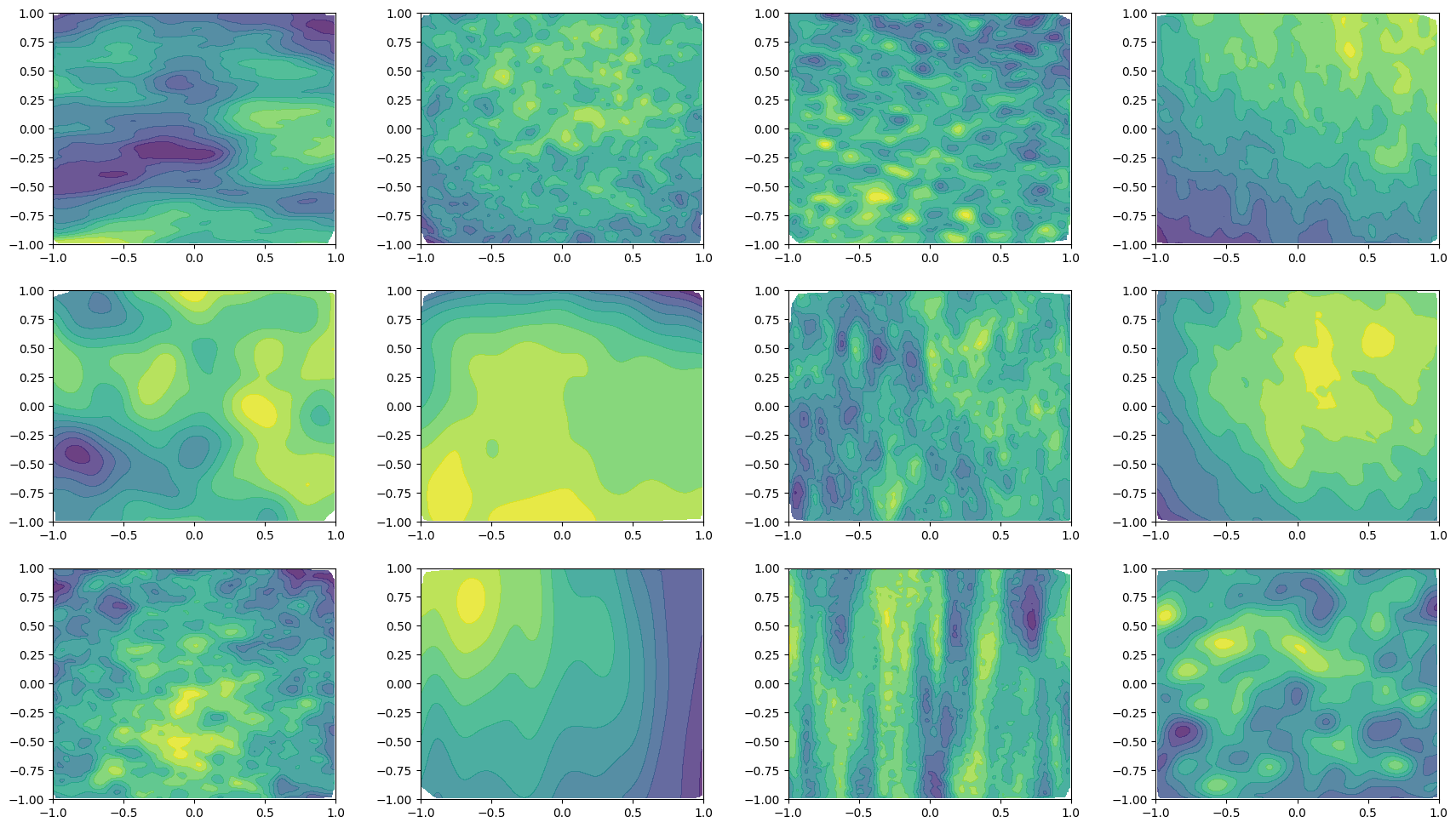}
    \caption{\textbf{Example of training samples used during pretraining with input dimension d=2}.}
    \label{fig:training_samples}
\end{figure}

Figure~\ref{fig:training_samples} shows 12 training sample tasks with $d=2$.

\subsection{Runtime Comparison}

Table \ref{tab:runtime_comparison} presents the runtime performance of all baselines evaluated on the Ackley function (\(q=64, H=10\)) across 15 independent runs. The results demonstrate that our proposed method achieves the fastest execution time.

\begin{table}[H]
    \centering
    \caption{Runtime comparison. Wall-clock time is measured over all $H=10$ iterations with $q=64$. The measurement unit is seconds.} 
    \label{tab:runtime_comparison}
    \small
    \begin{tabular}{@{}ccccc@{}}
    \toprule
    \textbf{Method} & $d=3$ & $d=5$ & $d=10$ & $d=15$ \\
    \midrule
    \textbf{FMTR} & $\mathbf{2.79 \pm 0.01}$ & $\mathbf{2.80 \pm 0.01}$ & $\mathbf{2.79 \pm 0.01}$ & $\mathbf{2.78 \pm 0.01}$ \\
    TuRBO & $8.15 \pm 0.67$ & $9.68 \pm 1.57$ & $12.17 \pm 1.37$ & $10.10 \pm 1.47$ \\
    DiBO & $180.35 \pm 3.89$ & $156.38 \pm 1.97$ & $205.71 \pm 3.28$ & $179.29 \pm 1.89$ \\
    BEEBO & $106.70 \pm 12.28$ & $198.17 \pm 42.50$ & $548.54 \pm 112.60$ & $789.35 \pm 149.53$ \\
    qLogEI & $93.69 \pm 16.69$ & $199.48 \pm 23.69$ & $350.10 \pm 56.05$ & $536.13 \pm 66.44$ \\
    qUCB & $56.85 \pm 8.69$ & $123.47 \pm 14.03$ & $436.70 \pm 31.04$ & $641.55 \pm 22.02$ \\
    qTS-Matheron & $3.77 \pm 0.22$ & $5.60 \pm 0.46$ & $12.88 \pm 4.95$ & $8.18 \pm 1.57$ \\
    GITBO & $24.79 \pm 0.10$ & $28.69 \pm 0.07$ & $36.79 \pm 2.01$ & $47.75 \pm 0.13$ \\
    \bottomrule
    \end{tabular}
\end{table}

\section{Additional pretraining details}
\label{app:pretraining_details}

\subsection{Synthetic objective prior}
\label{app:pretraining_prior}

This section specifies the two components of the synthetic objective in~\Eqref{eq:synthetic_objective}. For each optimization episode
\(\tau\), we sample a stationary kernel \(k_{\tau}\) and its hyperparameters.
Let \(p_{\tau}(\boldsymbol{\omega})\) denote the corresponding spectral
density. Following~\citet{rakotoarison2026}, we approximate a sample from
\(\mathcal{GP}(0,k_{\tau})\) using \(m\) random Fourier features:
\begin{equation}
    f_{\mathrm{RFF},\tau}(\mathbf{x})
    =
    \sqrt{\frac{2}{m}}\,
    \mathbf{z}_{\tau}^{\top}
    \cos\!\left(
        \mathbf{\Omega}_{\tau}\mathbf{x}
        +
        \mathbf{b}_{\tau}
    \right),
    \label{eq:appendix_rff}
\end{equation}
where
\[
    \mathbf{\Omega}_{\tau}
    =
    \begin{bmatrix}
        \boldsymbol{\omega}_{\tau,1}^{\top}\\
        \vdots\\
        \boldsymbol{\omega}_{\tau,m}^{\top}
    \end{bmatrix}
    \in
    \mathbb{R}^{m\times d_{\tau}},
    \qquad
    \boldsymbol{\omega}_{\tau,j}
    \overset{\mathrm{i.i.d.}}{\sim}
    p_{\tau}(\boldsymbol{\omega}).
\]
The feature coefficients and phase offsets are sampled as
\begin{equation}
    \mathbf{z}_{\tau}
    \sim
    \mathcal{N}(\mathbf{0},\mathbf{I}_{m}),
    \qquad
    \mathbf{b}_{\tau}
    \sim
    \mathcal{U}\!\left([0,2\pi]^{m}\right).
    \label{eq:appendix_rff_parameters}
\end{equation}
The cosine in~\Eqref{eq:appendix_rff} is applied element-wise.

Next, we add the randomized trend
\begin{equation}
    f_{\mathrm{trend},\tau}(\mathbf{x})
    =
    -a_{\tau}
    \left(
        \sum_{j=1}^{d_{\tau}}
        w_{\tau,j}
        \bigl(x_j-c_{\tau,j}\bigr)^2
    \right)^{p_{\tau}},
    \label{eq:appendix_trend}
\end{equation}
where \(a_{\tau}\), \(p_{\tau}\),
\(\mathbf{w}_{\tau}=(w_{\tau,1},\ldots,w_{\tau,d_{\tau}})\), and
\(\mathbf{c}_{\tau}=(c_{\tau,1},\ldots,c_{\tau,d_{\tau}})\) are sampled
independently for each episode. The resulting objective is
\[
    f_{\tau}(\mathbf{x})
    =
    f_{\mathrm{RFF},\tau}(\mathbf{x})
    +
    f_{\mathrm{trend},\tau}(\mathbf{x}).
\]
The explicit RFF representation makes each posterior path differentiable, so its maximizer can be obtained using gradient-based optimization. The kernel and its hyperparameters are sampled as part of the episode and are therefore available during target construction without task-specific estimation.

\subsection{Sampling kernel and trend function}
\label{app:kernel}

For each synthetic task, we sample the kernel and trend parameters as follows:
\begin{enumerate}
    \item \textbf{Kernel family.}
    We sample a Mat\'ern kernel with
    \(\nu\in\{1/2,3/2,5/2,\infty\}\), with probabilities proportional to
    \((0.1,1,1,1)\). The output scale is sampled from
    \(\mathcal{U}(0.1,1.0)\).

    \item \textbf{Lengthscale structure.}
    We sample \emph{isotropic}, \emph{ARD}, or \emph{grouped} lengthscales
    with probabilities \(0.2\), \(0.7\), and \(0.1\), respectively.
    Isotropic kernels use a shared lengthscale from
    \(\mathcal{U}(0.01,0.5)\) (our search domain being $\mathcal{X}_{\tau}:= [-1,1]^d)$, while ARD kernels sample each coordinate
    independently from the same distribution.
    For grouped kernels, dimensions are assigned to fast and slow groups
    with centers \(c_{\mathrm{fast}}<c_{\mathrm{slow}}\), which are
    independently perturbed across dimensions. This generates objectives
    with heterogeneous, potentially low effective dimensionality.
    A random rotation is additionally applied with probability
    \(p_{\mathrm{rot}}=0.3\) to generate non-axis-aligned functions.

    \item \textbf{Trend inclusion and location.}
    We omit the quadratic trend with probability
    \(p_{\mathrm{none}}=0.2\). Otherwise, its center
    \(\mathbf{c}_{\tau}\) is sampled from
    \(\mathcal{U}(-0.8,0.8)\) with probability
    \(p_{\mathrm{outside}}=0.25\), and from
    \(\mathcal{U}(-0.375,0.375)\) otherwise, allowing some optima to lie
    near the boundary of the search space.

    \item \textbf{Trend shape.}
    The weights are sampled as
    \(\mathbf{w}_{\tau}\sim
    \operatorname{exp}(\mathcal{U}(-0.8,0.8))\)
    and normalized by their mean, while
    \(p_{\tau}\) is sampled uniformly from \(\{0.5,1,2\}\).

    \item \textbf{Relative scale.}
    We choose \(a_{\tau}\) such that the variance ratio between
    \(f_{\mathrm{RFF},\tau}\) and \(f_{\mathrm{trend},\tau}\)
    lies in \([0.3,6.0]\), preventing either component from dominating
    the objective.
\end{enumerate}

For the other hyper-parameters, see Table~\ref{tab:pretraining_task_prior}.

\begin{table}[h!]
    \centering
    \caption{Hyperparameter settings.}
    \label{tab:pretraining_task_prior}
    \small
    \begin{tabular}{@{}ll@{}}
        \toprule
        \multicolumn{2}{c}{\textbf{Synthetic task prior hyperparameters}} \\
        \toprule
        Episode component & Distribution or value \\
        \midrule
        Search domain \(\mathcal{X}_{\tau}\) & $[-1,1]^d$\\
        Dimension \(d_{\tau}\) & $[2,3,..,14,15]$\\
        Batch size \(q_{\tau}\) & $[30,40,50,60,70,80]$\\
        Horizon \(H_{\tau}\) & $[5,6,7,8,9,10]$\\
        Initial context size \(n_{0,\tau}\) & $[10,20,30]$\\
        Observation noise \(\sigma\) & 0.01\\
        Number of RFF features \(m\) & 500d\\
        $T_{min}$                    & 0.4 \\
        $T_{max}$                    & 1.0 \\
        Learning rate Langevin dynamics $\eta$ & 0.05 \\
        Std of Langevin dynamics's noise term $\sigma_\mathrm{L}$ & 0.05 \\
        Base Langevin refinement steps $K$ & 7 \\
        Number of target location for each episode and history $M$                         & 200 \\
        $N_{roll}$                          & 1500 \\
        $T_{refresh}$                       & 2000 \\
        $\gamma$                            & 0.7 \\
        \toprule
        \multicolumn{2}{c}{\textbf{Model 
        hyperparameters}} \\
        \midrule
        Embedding hidden $d_{\mathrm{model}}$ & 128 \\
        Number of attention heads              & 4 \\
        Number of self-attention layers        & 6 \\
        Number of cross-attention layers $L_{dec}$ & 3 \\
        Feed-forward dimension                  & 128 \\
        Number of components for GMM $C$                                      & 5 \\
        Number of decoder layer for GMM          & 2 (Separate per component)  \\
        \toprule
        \multicolumn{2}{c}{\textbf{Training}} \\
        \midrule
        Optimizer                               & AdamW \\
        Learning rate                           & 5e-4 \\
        Weight decay                            & 1e-2 \\
        Number of warm up step                  & 50000 \\
        Number of burn in step                  & 37500 \\
        LR scheduler                            & Cosine annealing restart with polynomial \\
        Restart period for LR                   & 50000 \\
        Polynomial steps                        & 100000 \\
        Number of training steps                & 1000000 \\
        Batch size                               & 128 \\
        \toprule
        \multicolumn{2}{c}{\textbf{Evaluation}} \\
        \midrule
        Inference solver step                         & 100 \\
        
        \bottomrule
    \end{tabular}
\end{table}

\subsection{Closed-loop pretraining}
\label{app:closed_loop_training}

At DAgger rollout iteration \(s\), the probability of sampling the rollout batch from \(\tilde{p}_1\) is \begin{equation} \beta_s=\gamma^s, \label{eq:appendix_beta_schedule} \end{equation} where \(\gamma\in(0,1)\) controls the decay. The initial replay buffer is constructed from \(\tilde{p}_1\)-only rollouts, corresponding to \(s=0\) and \(\beta_0=1\), so that valid training examples are available before the first parameter update. Later rollouts increasingly use the current FMTR proposal as \(\beta_s\) decreases. For each visited state \((\mathcal{D}_r,h_r,H)\), we recompute \(\tilde{p}_1(\cdot\mid\mathcal{D}_r,h_r,H)\) and store \(M\) target locations together with the information required to evaluate the training objective. The complete rollout batch is then drawn from either \(\tilde{p}_1\) or the current FMTR proposal according to~\Eqref{eq:rollout_mixture}. Its observations are appended to the history before constructing the next state. In practice, for simplicity, we can use $p_1$ instead of $\tilde{p}_1$ for $B_r$ in Equation \ref{eq:rollout_mixture}, and store the samples from $p_1$ and their corresponding weights for computing the training loss in Equation~\ref{eq:training_objective}.

The replay buffer is partitioned into buckets
\(\{\mathcal{B}_n\}\), where \(\mathcal{B}_n\) contains states whose
histories have size \(n\). Let
\[
    \mathcal{S}_{\mathcal{B}}
    =
    \{n:\mathcal{B}_n\neq\varnothing\}
\]
denote the occupied context-size buckets. A training minibatch is sampled
by first drawing
\[
    n\sim\mathcal{U}(\mathcal{S}_{\mathcal{B}})
\]
and then drawing \(b\) examples uniformly from \(\mathcal{B}_n\). All
examples in the resulting minibatch have the same context length, which
also permits rectangular batching without padding contexts of different
sizes.

The buffer is augmented every \(T_{\mathrm{refresh}}\) parameter updates.
As \(s\) increases, the decreasing \(\beta_s\) exposes FMTR to histories
increasingly induced by its own proposals, while the target distribution
is recomputed at every visited state. Algorithm~\ref{alg:rollout_collection}
gives the corresponding rollout and replay procedures.

\begin{algorithm}[h!]
\caption{FMTR pretraining: full procedure}
\label{alg:rollout_collection}
\small
\begin{algorithmic}[1]
\Require Episode prior $p(\tau)$; update budget $N_{\mathrm{train}}$;
minibatch size $b$; targets per state $M$; episodes per collection
$N_{\mathrm{roll}}$; refresh period $T_{\mathrm{refresh}}$;
decay $\gamma\in(0,1)$

\State Initialize $\theta$, bucketed buffer $\mathcal{B}\leftarrow\varnothing$,
and counters $u\leftarrow0$, $s\leftarrow0$

\While{$u<N_{\mathrm{train}}$}
    \State $\beta_s\leftarrow\gamma^s$
    \Comment{\Eqref{eq:appendix_beta_schedule}}

    \For{$j=1,\ldots,N_{\mathrm{roll}}$}
        \State Sample $\tau\sim p(\tau)$;
        set $\mathcal{D}\leftarrow\mathcal{D}_0^\tau$

        \For{$r=0,\ldots,H_\tau-1$}
            \State $h\leftarrow H_\tau-r$, $\alpha\leftarrow h/H_\tau$

            \State Draw $M$ independent posterior paths
            using~\Eqref{eq:matheron_posterior_path}

            \State Maximize each path to obtain
            $\mathbf{x}_{\mathrm{TS}}^{(m)}$
            via~\Eqref{eq:posterior_path_maximization}

            \State Refine each $\mathbf{x}_{\mathrm{TS}}^{(m)}$
            for $K_{h,H_\tau}$ steps using
            \Eqref{eq:remaining_round_refinement},
            yielding $\mathbf{x}_1^{(m)}$

            \State $\mathcal{Z}\leftarrow
            \{(\mathbf{x}_1^{(m)},f_\tau(\mathbf{x}_1^{(m)}))\}_{m=1}^{M}$

            \State Append $(\tau,\mathcal{D},h,H_\tau,\mathcal{Z})$
            to $\mathcal{B}_{|\mathcal{D}|}$

            \State Draw $z\sim\operatorname{Bernoulli}(\beta_s)$
            \If{$z=1$}
                \State
                $B_r\sim
                p_1(\cdot\mid\mathcal{D},h,H_\tau)^{\otimes q_\tau}$
            \Else
                \State
                $B_r\sim
                p_\theta(\cdot\mid\mathcal{D},\alpha)^{\otimes q_\tau}$
            \EndIf

            \State Draw
            $\epsilon_{r,i}\overset{\mathrm{i.i.d.}}{\sim}
            \mathcal{N}(0,\sigma^2)$,
            $i=1,\ldots,q_\tau$

            \State Evaluate
            $y_{r,i}=f_\tau(\mathbf{x}_{r,i})+\epsilon_{r,i}$
            in parallel for $\mathbf{x}_{r,i}\in B_r$

            \State $\mathcal{D}\leftarrow
            \mathcal{D}\cup
            \{(\mathbf{x}_{r,i},y_{r,i})\}_{i=1}^{q_\tau}$
        \EndFor
    \EndFor

    \State $U\leftarrow\min(T_{\mathrm{refresh}},N_{\mathrm{train}}-u)$

    \State $\mathcal{S}_{\mathcal{B}}\leftarrow
    \{n:\mathcal{B}_n\neq\varnothing\}$

    \For{each of the next $U$ parameter updates}
        \State Draw
        $n\sim\mathcal{U}(\mathcal{S}_{\mathcal{B}})$

        \State Sample a minibatch $\mathcal{T}$ of $b$ records
        uniformly from $\mathcal{B}_n$

        \For{each record $(\tau,\mathcal{D},h,H,\mathcal{Z})\in\mathcal{T}$}
            \State Set $\alpha\leftarrow h/H$ and compute
            $w_{1:M}$ using~\Eqref{eq:target_weights}

            \State Draw $t\sim\mathcal{U}(0,1)$ and
            $\mathbf{x}_0^{(1:M)}
            \sim\mathcal{N}(\mathbf{0},\mathbf{I}_{d_\tau})^{\otimes M}$

        \State Set
        $\mathbf{x}_t^{(m)}
        \leftarrow
        (1-t)\mathbf{x}_0^{(m)}+t\mathbf{x}_1^{(m)}$
        and
        $\mathbf{u}_t^{(m)}
        \leftarrow
        \mathbf{x}_1^{(m)}-\mathbf{x}_0^{(m)}$,
        $m=1,\ldots,M$
        \EndFor

        \State Compute $\widehat{\mathcal{L}}$ by averaging
        the weighted losses in~\Eqref{eq:full_training_objective}
        over $\mathcal{T}$

        \State $\theta\leftarrow$
        \Call{OptimiserStep}{$\theta,\nabla_\theta\widehat{\mathcal{L}}$}

        \State $u\leftarrow u+1$
    \EndFor

    \State $s\leftarrow s+1$
\EndWhile

\State \Return $\theta$
\end{algorithmic}
\end{algorithm}

\begin{algorithm}[h!]
\caption{FMTR batch generation}
\label{alg:fmtr_inference}
\small
\begin{algorithmic}[1]
\Require GMM velocity network $v_{\theta}$; total horizon $H$, remaining horizon $h$; batch size $q$; ODE solver step $n$; context set $\mathcal{D}=\{(\mathbf{x}_i, y_i)\}_{i=1}^{n_c}$

\State $\alpha\leftarrow h/H$

\State $\mathbf{x}^{(1:q)} \sim \mathcal{N}(\mathbf{0}, \mathbf{I}_d)^{\otimes q}$

\For{$i=0,\ldots,n-1$}
    \State $t = i / n$

    \State $t_2 = (i+1) / n$

    \State $\mathbf{u}^{(k)}
\sim
v_{\theta}
\left(
\cdot
\mid
\mathbf{x}^{(k)}, t, \mathcal{D}, \alpha
\right),
\qquad
k=1,\ldots,q.$

    \State $\hat{\mathbf{x}}_1^{(1:q)} = \mathbf{x}^{(1:q)} + (1-t)\mathbf{u}^{(1:q)}$
    \Comment{Endpoint sample.}

    \State $\hat{\mathbf{x}}_0^{(1:q)}\sim \mathcal{N}(\mathbf{0}, \mathbf{I}_d)^{\otimes q}$
    \Comment{Re-noise source state.}

    \State $\mathbf{x}^{(1:q)}=(1-t_2)\hat{\mathbf{x}}_0^{(1:q)} + t_2 \hat{\mathbf{x}}_1^{(1:q)}$

\EndFor
\State \Return $\mathbf{x}^{(1:q)}$
\end{algorithmic}
\end{algorithm}

\section{Additional details on query target distribution construction}
\label{app:target_construction}

\subsection{Pathwise GP posterior sampling}
\label{app:pathwise_posterior}

Consider a history
\(\mathcal{D}=\{(\mathbf{x}_i,y_i)\}_{i=1}^{n_c}\), and define
\[
    \mathbf{X}
    =
    \begin{bmatrix}
        \mathbf{x}_1^{\top}\\
        \vdots\\
        \mathbf{x}_{n_c}^{\top}
    \end{bmatrix}
    \in
    \mathbb{R}^{n_c\times d_{\tau}},
    \qquad
    \mathbf{y}
    =
    (y_1,\ldots,y_{n_c})^{\top}.
\]
For episode \(\tau\), we treat
\(m_{\tau}(\mathbf{x})=f_{\mathrm{trend},\tau}(\mathbf{x})\) as the GP
mean and \(k_{\tau}\) as its covariance kernel. Let
\[
    \mathbf{K}_{\tau}
    =
    \bigl[
        k_{\tau}(\mathbf{x}_i,\mathbf{x}_j)
    \bigr]_{i,j=1}^{n_c},
    \qquad
    \mathbf{k}_{\tau,\mathbf{X}}(\mathbf{x})
    =
    \bigl(
        k_{\tau}(\mathbf{x},\mathbf{x}_1),
        \ldots,
        k_{\tau}(\mathbf{x},\mathbf{x}_{n_c})
    \bigr)^{\top},
\]
and let
\(\mathbf{m}_{\tau}
=
(m_{\tau}(\mathbf{x}_1),\ldots,m_{\tau}(\mathbf{x}_{n_c}))^{\top}\).

For posterior draw \(j\), we sample an independent zero-mean RFF prior path
\(g_{\tau}^{(j)}\), using the construction described in
Appendix~\ref{app:pretraining_prior} and
\citet{rahimi2007random}. We also draw
\[
    \widetilde{\boldsymbol{\epsilon}}^{(j)}
    \sim
    \mathcal{N}(
        \mathbf{0},
        \sigma^2\mathbf{I}_{n_c}
    ),
\]
where \(\sigma^2\) is the observation-noise variance, and define
\[
    \boldsymbol{\psi_{\tau}^{(j)}(\mathbf{x})}
    =
    \bigl(
        \psi_{\tau}^{(j)}(\mathbf{x}_1),
        \ldots,
        \psi_{\tau}^{(j)}(\mathbf{x}_{n_c})
    \bigr)^{\top}.
\]
Matheron's rule gives the correction coefficients
\begin{equation}
    \mathbf{v}^{(j)}
    =
    \left(
        \mathbf{K}_{\tau}
        +
        \sigma^2\mathbf{I}_{n_c}
    \right)^{-1}
    \left(
        \mathbf{y}
        -
        \mathbf{m}_{\tau}
        -
        \boldsymbol{\psi}_{\tau}^{(j)}
        -
        \widetilde{\boldsymbol{\epsilon}}^{(j)}
    \right),
    \label{eq:matheron_coefficients}
\end{equation}
and the corresponding posterior path is
\begin{equation}
    \widetilde{f}_{\tau}^{(j)}(\mathbf{x})
    =
    m_{\tau}(\mathbf{x})
    +
    \psi_{\tau}^{(j)}(\mathbf{x})
    +
    \mathbf{k}_{\tau,\mathbf{X}}(\mathbf{x})^{\top}
    \mathbf{v}^{(j)}.
    \label{eq:matheron_posterior_path}
\end{equation}
This separates the posterior sample into an RFF prior path and a
function-space correction determined by the observations
\citep{wilson2020efficiently}. If the pretraining implementation conditions
on noise-free observations,~\Eqref{eq:matheron_coefficients}
reduces to the case \(\sigma=0\), and the auxiliary noise draw is omitted.

For each of \(M\) independent posterior paths, we compute
\begin{equation}
    \mathbf{x}_{\mathrm{TS}}^{(j)}
    \in
    \arg\max_{\mathbf{x}\in\mathcal{X}_{\tau}}
    \widetilde{f}_{\tau}^{(j)}(\mathbf{x}),
    \qquad
    j=1,\ldots,M.
    \label{eq:posterior_path_maximization}
\end{equation}
The resulting empirical Thompson-sampling distribution is
\begin{equation}
    \widehat{p}_{\mathrm{TS}}
    (\cdot\mid\mathcal{D})
    =
    \frac{1}{M}
    \sum_{j=1}^{M}
    \delta_{\mathbf{x}_{\mathrm{TS}}^{(j)}}(\cdot),
    \label{eq:empirical_ts_distribution}
\end{equation}
where \(\delta_{\mathbf{x}}\) denotes a point mass at \(\mathbf{x}\).
The explicit RFF representation makes each posterior path differentiable,
so its maximizer can be obtained using gradient-based optimization. The
kernel and its hyperparameters are sampled as part of the episode and are
therefore available during target construction without task-specific estimation.

\subsection{Stochastic refinement of Thompson-sampling targets}
\label{app:horizon_refinement}

\textbf{Normalized updates.}
For a fixed episode and optimization state
\((\tau,\mathcal{D},h,H)\), the temperature and number of
refinement steps are given by
\Eqref{eq:remaining_round_temperature}. We define the normalized gradient as
\begin{equation}
    \mathbf{g}_\tau(\mathbf{x})
    =
    \begin{cases}
        \displaystyle
        \frac{\nabla f_\tau(\mathbf{x})}
             {\|\nabla f_\tau(\mathbf{x})\|_2},
        & \|\nabla f_\tau(\mathbf{x})\|_2>0,\\[1.2ex]
        \mathbf{0},
        & \text{otherwise}.
    \end{cases}
    \label{eq:normalized_refinement_gradient}
\end{equation}
Thus, the deterministic increment in
\Eqref{eq:remaining_round_refinement} has norm at most \(\eta\), independently of the objective-gradient magnitude. The temperature changes the number of updates through \(K_{h,H}\); it does not
rescale the gradient or noise within an individual update.

\textbf{Stochastic target construction.}
For each Thompson-sampling location
\(\mathbf{x}_{\mathrm{TS}}^{(j)}\), we initialize
\(\mathbf{x}^{(0,j)}=\mathbf{x}_{\mathrm{TS}}^{(j)}\)
and apply \Eqref{eq:remaining_round_refinement} using independent
perturbations
\(\{\boldsymbol{\xi}_k^{(j)}\}_{k=0}^{K_{h,H}-1}\).
Let
\[
    \Phi_{\tau,h,H}
    \left(
        \mathbf{x}_{\mathrm{TS}}^{(j)},
        \boldsymbol{\xi}_{0:K_{h,H}-1}^{(j)}
    \right)
    =
    \mathbf{x}^{(K_{h,H},j)}
\]
denote the resulting refinement map. The population target is
\begin{equation}
    p_1(\cdot\mid\mathcal{D},h,H)
    =
    (\Phi_{\tau,h,H})_{\#}
    \left[
        p_{\mathrm{TS}}(\cdot\mid\mathcal{D})
        \otimes
        \mathcal{N}(\mathbf{0},\mathbf{I}_{d_\tau})
        ^{\otimes K_{h,H}}
    \right].
    \label{eq:stochastic_target_pushforward}
\end{equation}
Including the perturbations in the source measure makes the
stochastic nature of the refinement explicit.
To prevent samples from leaving the domain, we apply reflection~\citep{lou2023reflected} after each refinement step whenever a point falls outside the domain bounds.
In practice, the
target is represented by \(M\) refined samples,
\begin{equation}
    \widehat{p}_1(\cdot\mid\mathcal{D},h,H)
    =
    \frac{1}{M}
    \sum_{j=1}^{M}
    \delta_{\mathbf{x}^{(K_{h,H},j)}}(\cdot).
    \label{eq:empirical_refined_target}
\end{equation}
These samples provide the target locations used in
Section~\ref{sec:model}; their relative contributions to training
are subsequently determined by the value weights in
\Eqref{eq:target_weights}.

\textbf{Finite-step interpretation.}
We use the distribution after exactly \(K_{h,H}\) updates,
rather than running the process to stationarity. The construction
therefore retains the Thompson-sampling initialization as part of
the target definition.
Accordingly, the target is defined by the finite-step refinement dynamics rather than by a stationary Langevin distribution.
The values of
\(K,\eta,\sigma_{\mathrm{L}},T_{\min}\), and \(T_{\max}\)
are reported in Table~\ref{tab:pretraining_task_prior}.

\textbf{Scope of the refinement procedure.}
The target construction only requires samples from a finite
refinement procedure, so other procedures could be considered
for different synthetic-task families. Related samplers address
departures from smooth continuous objectives:
P-LMC~\citep{chatterji2020langevin} treats nonsmooth potentials
under assumptions that include subgradient access, whereas
DLP~\citep{zhang2022langevin} uses gradients of a differentiable
extension to construct proposals over discrete variables.

\subsection{Value-weighted target distribution}
\label{app:value_weighting}

For a fixed episode \(\tau\) and state \((\mathcal{D},h,H)\), write
\(T=T(h,H)>0\) and suppress the conditioning of \(p_1\) and
\(\tilde{p}_1\).
Refinement changes query locations using the objective gradient, but
the resulting basin masses also depend on the Thompson-sampling
initialization and finite-step dynamics.
Value weighting additionally adjusts these masses according to the
attained objective values.

\textbf{Gibbs variational characterization.}
Consider
\begin{equation}
    \tilde{p}_1
    =
    \operatorname*{argmax}_{p\ll p_1}
    \left\{
        \mathbb{E}_{p}[f_\tau(\mathbf{x})]
        -
        T\operatorname{KL}(p\|p_1)
    \right\},
    \label{eq:tilt_problem}
\end{equation}
where \(p\ll p_1\) denotes absolute continuity, and we restrict
to probability distributions with finite objective expectations
and KL divergences.
Define
\begin{equation}
    Z
    =
    \mathbb{E}_{p_1}
    \left[\exp\left(\frac{f_\tau(\mathbf{x})}{T}\right)\right],
    \qquad
    \tilde{p}_1(\mathrm{d}\mathbf{x})
    =
    \frac{\exp(f_\tau(\mathbf{x})/T)}{Z}
    p_1(\mathrm{d}\mathbf{x}).
    \label{eq:normalized_tilt}
\end{equation}
Assume \(0<Z<\infty\) and
\(\mathbb{E}_{\tilde{p}_1}[|f_\tau(\mathbf{x})|]<\infty\).
For any admissible \(p\), substituting the density ratio from
\Eqref{eq:normalized_tilt} gives
\begin{equation}
    \mathbb{E}_{p}[f_\tau(\mathbf{x})]
    -
    T\operatorname{KL}(p\|p_1)
    =
    T\log Z
    -
    T\operatorname{KL}(p\|\tilde{p}_1).
    \label{eq:gibbs_variational_identity}
\end{equation}
Nonnegativity of KL divergence establishes that \(\tilde{p}_1\)
is the unique maximizer.

\textbf{Relation to refinement.}
For a fixed reference distribution \(p_1\), reducing \(T\) increases
the relative weight assigned to larger objective values.
In FMTR, \(T\) also changes the reference distribution through the
refinement budget \(K_{h,H}\).
The Gibbs characterization applies to this finite-step reference law
and requires no stationary-distribution assumption for the refinement
dynamics.
Moreover, \(\tilde{p}_1\ll p_1\): value weighting redistributes
existing probability mass but cannot recover regions assigned zero
mass by \(p_1\).
Appendix~\ref{app:weighted_training} explains its implementation
through weighted training.

\section{Model architecture and training details}
\label{app:model_details}

\begin{figure}[h] 
    \centering   \includegraphics[width=0.8\linewidth]{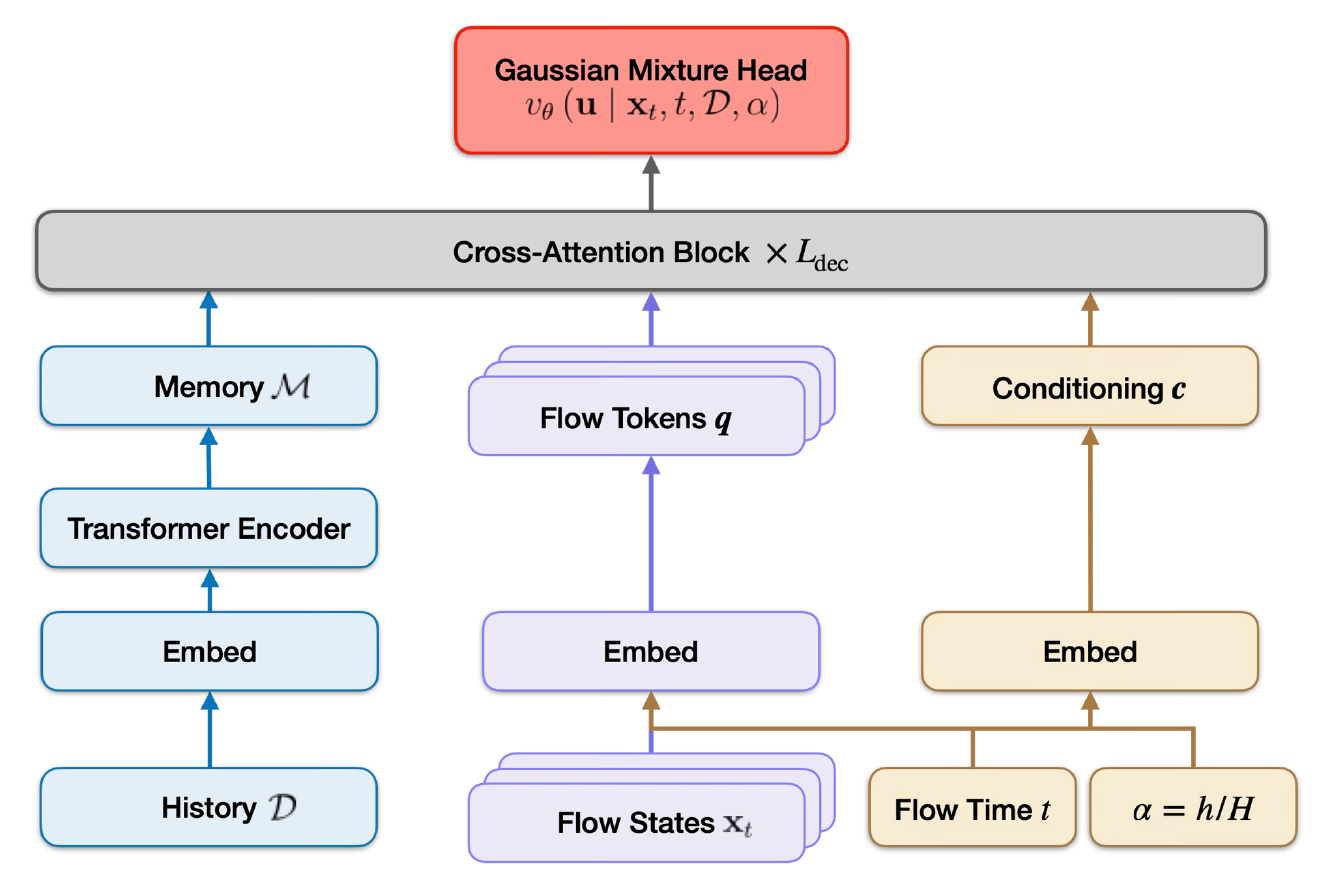}
    \caption{\textbf{Model Architecture}.}
    \label{fig:model_architecture}
\end{figure}

This section provides additional details for the conditional proposal model
introduced in Section~\ref{sec:model}. We first specify the input
representations and conditional decoder, then expand the weighted training
objective in~\Eqref{eq:training_objective}, and finally give
additional details on batch generation. Model architecture figure is given in Figure \ref{fig:model_architecture}.

\subsection{Input representation and conditional decoder}
\label{app:model_architecture}

\textbf{Variable-dimensional inputs.}
To support objectives with input dimension \(d\leq D\), where \(D\) is the
maximum dimension supported by FMTR, each location is zero-padded to \(D\)
coordinates and paired with an active-dimension mask
\(\mathbf{m}\in\{0,1\}^{D}\). Padded coordinates remain fixed at zero
throughout generation. Section~\ref{sec:model} uses the active
\(d\)-dimensional coordinates for readability.

\textbf{Context encoding.}
The location and objective value of each observation are embedded as
\begin{equation}
    \mathbf{z}_i
    =
    E_x(\mathbf{x}_i)+E_y(y_i),
    \label{eq:context_token}
\end{equation}
with \(\mathbf{z}_i\in\mathbb{R}^{d_{\mathrm{model}}}\).
A pre-normalized transformer encoder processes these tokens together with
a learned global token~\citep{vaswani2017attention}. We retain the
contextualized observation tokens as the memory
\[
    \mathcal{M}
    \in
    \mathbb{R}^{n_c\times d_{\mathrm{model}}},
\]
rather than pooling them into a single representation.

No positional encodings are used for the observations. Permuting the
history therefore only permutes the corresponding memory tokens, leaving
cross-attention to the memory unchanged. Since \(\mathcal{M}\) depends on
the observed history but not on the flow state, it is computed once per
optimization round and reused throughout flow generation.

\textbf{Flow-state embedding.}
The semantics of \(\mathbf{x}_t\) vary with flow time: near \(t=0\), it is
dominated by the Gaussian base sample, while near \(t=1\), it approaches a
query location. We therefore use
\begin{equation}
    \mathbf{q}
    =
    E_t(\mathbf{x}_t,t, \alpha)
    =
    E_x(\mathbf{x}_t)
    +
    (1-t)\operatorname{MLP}(\mathbf{x}_t,t,\alpha),
    \label{eq:target_embed}
\end{equation}
where both terms lie in
\(\mathbb{R}^{d_{\mathrm{model}}}\).
The additional term vanishes at \(t=1\), giving
\(E_t(\mathbf{x},1, \alpha)=E_x(\mathbf{x})\).

\textbf{Conditional decoder with AdaLN-Zero.}
The flow token \(\mathbf{q}\) is processed by \(L_{\mathrm{dec}}\)
cross-attention blocks, each attending to \(\mathcal{M}\) without
self-attention among flow tokens. Following AdaLN-Zero from
DiT~\citep{peebles2023scalable}, flow time and the remaining-round fraction
are combined as
\begin{equation}
    \mathbf{c}
    =
    E_{\mathrm{time}}(t)
    +
    E_{\mathrm{horizon}}(\alpha),
    \label{eq:decoder_conditioning}
\end{equation}
where \(E_{\mathrm{time}}\) uses random Fourier features of \(t\) and
\(E_{\mathrm{horizon}}\) is a multilayer perceptron.

Each decoder block maps this conditioning vector to scale, shift, and
gating parameters,
\begin{equation}
    \left(
        \boldsymbol{\gamma}_1,\boldsymbol{\beta}_1,\mathbf{g}_1,
        \boldsymbol{\gamma}_2,\boldsymbol{\beta}_2,\mathbf{g}_2
    \right)
    =
    W_{\mathrm{ada}}\operatorname{SiLU}(\mathbf{c}),
    \label{eq:adaln_parameters}
\end{equation}
and applies
\begin{align}
    \mathbf{q}
    &\leftarrow
    \mathbf{q}
    +
    \mathbf{g}_1\odot
    \operatorname{MHA}
    \left(
        (1+\boldsymbol{\gamma}_1)\odot
        \operatorname{LN}(\mathbf{q})+\boldsymbol{\beta}_1,
        \operatorname{LN}(\mathcal{M})
    \right),
    \label{eq:decoder_attention}
    \\
    \mathbf{q}
    &\leftarrow
    \mathbf{q}
    +
    \mathbf{g}_2\odot
    \operatorname{FFN}
    \left(
        (1+\boldsymbol{\gamma}_2)\odot
        \operatorname{LN}(\mathbf{q})+\boldsymbol{\beta}_2
    \right).
    \label{eq:decoder_feedforward}
\end{align}
Here, the second argument of \(\operatorname{MHA}\) supplies its keys and
values. Following AdaLN-Zero, \(W_{\mathrm{ada}}\) is initialized to zero,
so the residual gates vanish at initialization and each decoder block
initially acts as the identity on its query token.
The context memory is normalized but not modulated by the flow-time
conditioning, since it represents fixed observed data.

\textbf{Gaussian mixture head.}
The final decoder representation is projected to the parameters of the
velocity distribution in~\Eqref{eq:gmm}: \(C\) mixture logits,
\(C\) mean vectors, and \(C\) diagonal scale vectors. The mixture weights
are obtained by applying a softmax to the logits. The predicted scales are
lower-bounded by a small positive constant for numerical stability.

\subsection{Training objective}
\label{app:weighted_training}

We now expand the objective in~\Eqref{eq:training_objective}.
Let
\(\rho_{\mathrm{train}}(\tau,\mathcal{D},h,H)\)
denote the distribution over training records produced by the closed-loop
collection and context-length-stratified sampling procedure described in
Section~\ref{sec:closed_loop_training}.
For a sampled record, define
\(\alpha=h/H\) and consider the population flow-matching objective under
the tilted target,
\begin{equation}
    \mathcal{L}_{\mathrm{tilt}}(\theta)
    =
    -\mathbb{E}_{(\tau,\mathcal{D},h,H)\sim\rho_{\mathrm{train}}}
    \left[
        \mathbb{E}_{\substack{
            t\sim\mathcal{U}(0,1),\\
            \tilde{\mathbf{x}}_1
            \sim
            \tilde p_1(\cdot\mid\mathcal{D},h,H),\\
            \mathbf{x}_0
            \sim
            p_0
        }}
        \left[
            \log
            v_\theta
            \left(
                \tilde{\mathbf{u}}_t
                \mid
                \tilde{\mathbf{x}}_t,
                t,
                \mathcal{D},
                \alpha
            \right)
        \right]
    \right],
    \qquad
    \alpha=\frac{h}{H},
    \label{eq:tilted_population_objective}
\end{equation}
where
\(p_0=\mathcal{N}(\mathbf{0},\mathbf{I}_d)\),
\(\tilde{\mathbf{x}}_t=(1-t)\mathbf{x}_0+t\tilde{\mathbf{x}}_1\), and
\(\tilde{\mathbf{u}}_t=\tilde{\mathbf{x}}_1-\mathbf{x}_0\).

Direct samples from \(\tilde p_1\) are not available in our target
construction. Instead, we have samples from
\(p_1(\cdot\mid\mathcal{D},h,H)\).
From~\Eqref{eq:tilted_refined_target_distribution},
\begin{equation}
    \tilde p_1(\mathbf{x}\mid\mathcal{D},h,H)
    =
    \frac{
        p_1(\mathbf{x}\mid\mathcal{D},h,H)
        \exp\!\left(f_\tau(\mathbf{x})/T(h,H)\right)
    }{
        Z(\tau,\mathcal{D},h,H)
    },
    \label{eq:tilted_target_normalized}
\end{equation}
where
\begin{equation}
    Z(\tau,\mathcal{D},h,H)
    =
    \mathbb{E}_{\mathbf{x}\sim p_1(\cdot\mid\mathcal{D},h,H)}
    \left[
        \exp\!\left(f_\tau(\mathbf{x})/T(h,H)\right)
    \right].
    \label{eq:tilted_normalizer}
\end{equation}
Hence, for any \(\tilde p_1\)-integrable function \(\psi\), and assuming $0 < Z < +\infty$
\begin{align}
    \mathbb{E}_{\tilde p_1}
    [\psi(\mathbf{x}_1)]
    &=
    \int
    \psi(\mathbf{x}_1)
    \tilde p_1(\mathbf{x}_1\mid\mathcal{D},h,H)
    \,\mathrm{d}\mathbf{x}_1
    \nonumber\\
    &=
    \frac{1}{Z}
    \int
    \psi(\mathbf{x}_1)
    p_1(\mathbf{x}_1\mid\mathcal{D},h,H)
    \exp\!\left(
        \frac{f_\tau(\mathbf{x}_1)}{T(h,H)}
    \right)
    \,\mathrm{d}\mathbf{x}_1
    \nonumber\\
    &=
    \frac{
        \mathbb{E}_{p_1}
        \left[
            \exp\!\left(
                f_\tau(\mathbf{x}_1)/T(h,H)
            \right)
            \psi(\mathbf{x}_1)
        \right]
    }{
        \mathbb{E}_{p_1}
        \left[
            \exp\!\left(
                f_\tau(\mathbf{x}_1)/T(h,H)
            \right)
        \right]
    }.
    \label{eq:tilted_expectation_exact}
\end{align}

For \(M\) independent samples
\(\mathbf{x}_1^{(i)}
\sim p_1(\cdot\mid\mathcal{D},h,H)\),
define the unnormalized importance weights
\begin{equation}
    s_i
    =
    \exp\!\left(
        \frac{f_\tau(\mathbf{x}_1^{(i)})}{T(h,H)}
    \right),
    \qquad i=1,\ldots,M.
    \label{eq:unnormalized_target_weights}
\end{equation}
Self-normalized importance sampling
\citep{aeckerle2026error} then approximates
\Eqref{eq:tilted_expectation_exact} as
\begin{equation}
    \mathbb{E}_{\tilde p_1}
    [\psi(\mathbf{x}_1)]
    \approx
    \frac{
        \sum_{i=1}^{M}
        s_i\psi(\mathbf{x}_1^{(i)})
    }{
        \sum_{j=1}^{M}s_j
    }
    =
    \sum_{i=1}^{M}
    w_i\psi(\mathbf{x}_1^{(i)}),
    \label{eq:tilted_expectation_snis}
\end{equation}
where
\begin{align}
    w_i
    &=
    \frac{s_i}{\sum_{j=1}^{M}s_j}
    \nonumber\\
    &=
    \frac{
        \exp\!\left(
            f_\tau(\mathbf{x}_1^{(i)})/T(h,H)
        \right)
    }{
        \sum_{j=1}^{M}
        \exp\!\left(
            f_\tau(\mathbf{x}_1^{(j)})/T(h,H)
        \right)
    }
    \nonumber\\
    &=
    \operatorname{softmax}_i
    \left(
        \frac{f_\tau(\mathbf{x}_1^{(i)})}{T(h,H)}
    \right).
    \label{eq:target_weights}
\end{align}
Equivalently, using the density ratio from
\Eqref{eq:tilted_target_normalized},
\begin{equation}
    \frac{
        \tilde p_1(\mathbf{x}_1^{(i)}
        \mid\mathcal{D},h,H)
    }{
        p_1(\mathbf{x}_1^{(i)}
        \mid\mathcal{D},h,H)
    }
    =
    \frac{s_i}{Z},
\end{equation}
so that the unknown normalizing constant \(Z\) cancels after
normalizing the importance weights:
\begin{equation}
    \frac{
        \tilde p_1(\mathbf{x}_1^{(i)})/
        p_1(\mathbf{x}_1^{(i)})
    }{
        \sum_{j=1}^{M}
        \tilde p_1(\mathbf{x}_1^{(j)})/
        p_1(\mathbf{x}_1^{(j)})
    }
    =
    \frac{s_i}{\sum_{j=1}^{M}s_j}
    =
    w_i.
    \label{eq:density_ratio_weights}
\end{equation}

We now apply this approximation to the inner expectation in
\Eqref{eq:tilted_population_objective}. For
\[
    \psi(\mathbf{x}_1)
    =
    \mathbb{E}_{\substack{
        t\sim\mathcal{U}(0,1),\\
        \mathbf{x}_0\sim p_0
    }}
    \left[
        \log
        v_\theta
        \left(
            \mathbf{u}_t
            \mid
            \mathbf{x}_t,
            t,
            \mathcal{D},
            \alpha
        \right)
    \right],
\]
where
\(\mathbf{x}_t=(1-t)\mathbf{x}_0+t\mathbf{x}_1\)
and
\(\mathbf{u}_t=\mathbf{x}_1-\mathbf{x}_0\),
\Eqref{eq:tilted_expectation_snis} gives
\begin{equation}
\begin{aligned}
&\mathbb{E}_{\substack{
    t\sim\mathcal{U}(0,1),\\
    \tilde{\mathbf{x}}_1\sim
    \tilde p_1(\cdot\mid\mathcal{D},h,H),\\
    \mathbf{x}_0\sim p_0
}}
\left[
    \log
    v_\theta
    \left(
        \tilde{\mathbf{u}}_t
        \mid
        \tilde{\mathbf{x}}_t,
        t,
        \mathcal{D},
        \alpha
    \right)
\right]
\\
&\qquad\approx
\mathbb{E}_{\substack{
    t\sim\mathcal{U}(0,1),\\
    \mathbf{x}_1^{(1:M)}
    \sim p_1(\cdot\mid\mathcal{D},h,H)^{\otimes M},\\
    \mathbf{x}_0^{(1:M)}
    \sim p_0^{\otimes M}
}}
\left[
    \sum_{i=1}^{M}
    w_i
    \log
    v_\theta
    \left(
        \mathbf{u}_t^{(i)}
        \mid
        \mathbf{x}_t^{(i)},
        t,
        \mathcal{D},
        \alpha
    \right)
\right].
\end{aligned}
\label{eq:inner_objective_snis}
\end{equation}
with
\[
    \mathbf{x}_t^{(i)}
    =
    (1-t)\mathbf{x}_0^{(i)}
    +
    t\mathbf{x}_1^{(i)},
    \qquad
    \mathbf{u}_t^{(i)}
    =
    \mathbf{x}_1^{(i)}-\mathbf{x}_0^{(i)}.
\]
Combining this approximation with the outer expectation over training
records gives the practical training objective
\begin{equation}
\begin{aligned}
    \mathcal{L}(\theta)
    =
    -\mathbb{E}_{\substack{
        (\tau,\mathcal{D},h,H)\sim\rho_{\mathrm{train}},\\
        t\sim\mathcal{U}(0,1),\\
        \mathbf{x}_1^{(1:M)}
        \sim
        p_1(\cdot\mid\mathcal{D},h,H)^{\otimes M},\\
        \mathbf{x}_0^{(1:M)}
        \sim
        \mathcal{N}(\mathbf{0},\mathbf{I}_d)^{\otimes M}
    }}
    \left[
        \sum_{i=1}^{M}
        w_i
        \log
        v_{\theta}
        \left(
            \mathbf{u}_t^{(i)}
            \mid
            \mathbf{x}_t^{(i)},
            t,
            \mathcal{D},
            \alpha
        \right)
    \right],
    \qquad
    \alpha=\frac{h}{H}.
\end{aligned}
\label{eq:full_training_objective}
\end{equation}
This is the objective summarized in
\Eqref{eq:training_objective}.
For a fixed target pool, the weighted loss corresponds exactly to the
empirical distribution
\begin{equation}
    \hat{\tilde p}_{1,M}
    =
    \sum_{i=1}^{M}
    w_i\delta_{\mathbf{x}_1^{(i)}}.
\end{equation}
As an estimator of the population expectation under \(\tilde p_1\),
self-normalized importance sampling is generally biased for finite \(M\),
but is consistent under standard integrability conditions.

The target construction and value weighting use the synthetic objective in different ways. The refinement in
Section~\ref{sec:target_distribution} moves Thompson-sampling locations to define \(p_1\), whereas~\Eqref{eq:target_weights} reweights samples from \(p_1\) to approximate training under \(\tilde p_1\). The objective values are used only for pretraining supervision and are not additional inputs to
\(p_\theta(\mathbf{x}\mid\mathcal{D},\alpha)\).

\subsection{Computational Resources}

We trained FMTR on one NVIDIA H100 80GB HBM3 GPU. Training took approximately 19 hours per run. All models are evaluated on Tesla
V100-SXM2-32GB GPUs.

\section{Experimental details}
\label{app:experimental_details}

This section supplements Section~\ref{sec:experiments} with descriptions
of the baseline methods. FMTR's pretraining procedure and architecture are
given in Appendices~\ref{app:pretraining_details} and~\ref{app:model_details}. For evaluation, since FMTR is trained on the normalized domain $\mathcal{X}=[-1,1]^d$, the observed query points are first mapped to $\mathcal{X}$. FMTR then generates new query points within $\mathcal{X}$, which are subsequently mapped back to the original domain for function evaluation.

\subsection{Baselines}
\label{app:baselines}

\textbf{TuRBO}~\citep{eriksson2019scalable}.
Trust Region Bayesian Optimization fits local GP models within adaptive trust regions.
Their sizes expand after consecutive improvements and contract after unsuccessful evaluations, with
restarts when a region becomes too small.
Its Thompson-sampling batch rule selects candidates within these restricted regions, and can allocate evaluations across multiple local models.
TuRBO therefore provides a baseline based on local modeling
and adaptive search regions, while retaining task-specific
surrogate fitting and online candidate selection. In our experiments, we specify 5 trust regions for TuRBO.

\textbf{DiBO}~\citep{yun2025posterior}.
This method fits a diffusion model to evaluated designs and an ensemble
of surrogate models to their objective values. It then adapts the diffusion
sampler through relative trajectory balance toward a target combining the
learned design prior with an ensemble-based UCB score. Generated candidates
undergo local search and filtering before a batch is evaluated. Model
fitting and sampler adaptation are repeated as observations accumulate,
providing a generative baseline trained on the current optimization task.
We use the authors' \href{https://github.com/umkiyoung/DiBO}{official implementation}.

\textbf{GIT-BO}~\citep{yu2026git}.
GIT-BO uses a pretrained TabPFN surrogate to predict objective values and
uncertainty from the observed history. Gradients of the predictive mean
identify a low-dimensional subspace in which candidate locations are
sampled and scored using a UCB-based acquisition. For the batch setting,
we use the native implementation provided in the authors'
\href{https://github.com/rosenyu304/GITBO}{official repository}, setting
\(N_{\mathrm{CANDIDATES}}=q\). The method constructs \(q\) candidate pools,
evaluates the acquisition independently for each batch slot, and selects
the maximizing candidate from each pool. The batch is therefore proposed
without intermediate observations, but its members are not coupled through
a joint acquisition or within-batch conditioning.

\textbf{BEEBO}~\citep{teufel2024}.
Batched Energy--Entropy Bayesian Optimization uses a GP-based acquisition
that combines predicted objective values with the information gained from
evaluating a batch. A temperature controls the exploration--exploitation
trade-off, while the information term accounts for dependencies through
the joint posterior covariance. Candidate locations are optimized jointly,
providing a batch-native acquisition baseline that retains online GP
fitting and numerical batch selection.
We use the BoTorch implementation~\citep{balandat2020botorch} with the default settings: \texttt{meanBEEBO} variant, keeping the temperature fixed at $T=2$
and switching to $T=0$ (full exploitation) in the final round.

\textbf{qLogEI, qTS-Matheron, and qUCB.}
These baselines use a GP surrogate with different batch-selection rules and
are implemented with BoTorch~\citep{balandat2020botorch}.
qLogEI uses a numerically stabilized, smoothed log-space formulation of
batch expected improvement~\citep{ament2023unexpected}.
qTS-Matheron draws \(q\) independent posterior paths through Matheron's rule
and maximizes each path to obtain one query~\citep{kandasamyparallelts,wilson2020efficiently}.
qUCB extends upper confidence bounds to batches through a Monte Carlo
acquisition based on the joint posterior~\citep{wilson2018maximizing}.
Thus, qLogEI and qUCB optimize batch acquisitions, whereas qTS-Matheron
constructs the batch through independent posterior-path optimizations.

For all baselines that optimize acquisition using BoTorch, we set \texttt{raw\_samples} = 512 and \texttt{num\_restarts} = 128.

\subsection{Synthetic Benchmark Functions}
\label{app:synthetic_tasks}

We evaluate our method on six standard synthetic benchmark functions with different optimization landscapes. The Ackley, Levy, and Rastrigin functions are highly multimodal, while Rosenbrock, Dixon--Price, and Styblinski--Tang provide complementary nonconvex landscapes. We define the functions below, where $\mathbf{x}=(x_1,\ldots,x_d)$ and $d$ denotes the input dimensionality.

\paragraph{Ackley.}
\begin{equation}
f(\mathbf{x})
=
-20\exp\left(
-0.2\sqrt{\frac{1}{d}\sum_{i=1}^{d}x_i^2}
\right)
-
\exp\left(
\frac{1}{d}\sum_{i=1}^{d}\cos(2\pi x_i)
\right)
+20+e,
\end{equation}
with domain
\[
x_i\in[-32.768,32.768],
\qquad i=1,\ldots,d.
\]

\paragraph{Levy.}
Let
\[
w_i=1+\frac{x_i-1}{4}.
\]
The Levy function is
\begin{equation}
f(\mathbf{x})
=
\sin^2(\pi w_1)
+
\sum_{i=1}^{d-1}
(w_i-1)^2
\left[
1+10\sin^2(\pi w_i+1)
\right]
+
(w_d-1)^2
\left[
1+\sin^2(2\pi w_d)
\right],
\end{equation}
with domain
\[
x_i\in[-10,10],
\qquad i=1,\ldots,d.
\]

\paragraph{Rosenbrock.}
\begin{equation}
f(\mathbf{x})
=
\sum_{i=1}^{d-1}
\left[
100(x_{i+1}-x_i^2)^2+(x_i-1)^2
\right],
\end{equation}
with domain
\[
x_i\in[-5,10],
\qquad i=1,\ldots,d.
\]

\paragraph{Rastrigin.}
\begin{equation}
f(\mathbf{x})
=
10d
+
\sum_{i=1}^{d}
\left[
x_i^2-10\cos(2\pi x_i)
\right],
\end{equation}
with domain
\[
x_i\in[-5.12,5.12],
\qquad i=1,\ldots,d.
\]

\paragraph{Dixon--Price.}
\begin{equation}
f(\mathbf{x})
=
(x_1-1)^2
+
\sum_{i=2}^{d}
i(2x_i^2-x_{i-1})^2,
\end{equation}
with domain
\[
x_i\in[-10,10],
\qquad i=1,\ldots,d.
\]

\paragraph{Styblinski--Tang.}
\begin{equation}
f(\mathbf{x})
=
\frac{1}{2}
\sum_{i=1}^{d}
\left(
x_i^4-16x_i^2+5x_i
\right),
\end{equation}
with domain
\[
x_i\in[-5,5],
\qquad i=1,\ldots,d.
\]

\subsection{Real-world benchmark tasks}
\label{app:realworld_tasks}

We consider four data-derived optimization benchmarks described
by \citet{hsieh2021reinforced}.

\textbf{Oil-well depth (oil-4D).}
This task uses the New York State Oil, Gas, and Other Regulated
Wells dataset. The four inputs are the latitude and longitude
of the surface and bottom of each well, and the objective
is to identify wells with the greatest drilled depth.

\textbf{XGBoost hyperparameter optimization (hpobenchXGB-6D).}
Precomputed HPOBench evaluations associate six XGBoost
hyperparameters with classification performance. These include
learning rate, \(L_1\) and \(L_2\) regularization, and subsampling
ratios. The objective favors configurations with higher
predictive performance.

\textbf{Asteroid diameter (Asteroid-12D).}
Each candidate is represented by twelve numerical asteroid
descriptors. The task seeks candidates associated with the
largest diameters.

\textbf{Air quality (pm25-14D).}
This benchmark uses air-quality observations collected in
northern Taiwan in 2015. Its fourteen inputs include sulfur
dioxide, carbon monoxide, and ozone measurements.
The objective is to identify conditions associated with
the highest \(\mathrm{PM}_{2.5}\) concentration.

Since these tasks are defined on a discrete set of points, we interpolate them onto a continuous domain using a radial basis function (RBF) interpolator with a linear kernel and 32 nearest neighbors. We implement the interpolation using the \texttt{RBFInterpolator} provided by SciPy. Further details are available in the \href{https://docs.scipy.org/doc/scipy/reference/generated/scipy.interpolate.RBFInterpolator.html}{SciPy documentation}.

\end{document}